\pdfoutput=1

\documentclass[11pt]{article}

\usepackage[]{acl}

\usepackage{times}
\usepackage{latexsym}

\usepackage[T1]{fontenc}

\usepackage[utf8]{inputenc}

\usepackage{microtype}

\usepackage{inconsolata}

\usepackage{amssymb, amsmath}
\usepackage{siunitx}
\usepackage{bm}
\usepackage{graphicx}
\usepackage{booktabs}
\usepackage{tabularx}
\usepackage{multirow}
\usepackage{adjustbox}
\usepackage{array}
\usepackage{pifont}
\usepackage{xspace}
\usepackage{tcolorbox}
\usepackage{colortbl}
\usepackage{hyperref}
\usepackage{arydshln}
\usepackage{wrapfig}

\usepackage{makecell}
\usepackage{color, xcolor}
\usepackage{pgfplots}
\definecolor{tiffanyblue}{RGB}{129,216,208}
\definecolor{bangdiblue}{RGB}{0,149,182}
\definecolor{kleinblue}{RGB}{0,47,167}
\usepackage{amssymb}
\usepackage{circledtext}
\usetikzlibrary{shapes}
\usetikzlibrary{arrows,decorations.pathmorphing,backgrounds,positioning,fit,petri}
\usetikzlibrary{arrows.meta,fit,shapes.arrows}
\usetikzlibrary{positioning,shadows,patterns}
\usepackage{caption} 
\usepackage{tikz}
\usepackage{subfig}
\usepackage{tkz-kiviat,pgfplots}
\pgfplotsset{compat=newest}

\definecolor{orange}{HTML}{ff9900} 

\definecolor{green}{HTML}{34a853}
\definecolor{lightgreen}{HTML}{b6d7a8}
\definecolor{seagreen}{HTML}{3CB371}

\definecolor{lightgray1}{HTML}{d9d9d9}
\definecolor{lightyellow2}{HTML}{ffe599}

\definecolor{blue}{HTML}{4285f4}
\definecolor{lightblue}{HTML}{9fc5e8}
\definecolor{lightcornflowerblue3}{HTML}{c9daf8}

\definecolor{purple}{HTML}{9900ff} 
\definecolor{lightpurple1}{HTML}{8e7cc3}
\definecolor{lightpurple}{HTML}{b4a7d6}
\definecolor{lightred}{HTML}{e06666}

\usepackage{soul}
\colorlet{exqcolor}{lightgreen!70}
\colorlet{exccolor}{lightgray1!55}
\colorlet{exdcolor}{lightyellow2!90}
\colorlet{exacolor}{lightblue!90}
\colorlet{execolor}{lightred!50}
\colorlet{exncolor}{lightpurple!50}

\newcommand{\ours}{\textsc{Front}\xspace}
\newcommand{\headercolor}{\rowcolor{gray!5}}
\newcommand{\headercolorlong}{\rowcolor{gray!17}}
\newcommand{\vani}{{\textsc{Vanilla}}}

\newcommand{\eg}{\emph{e.g.}}
\makeatletter
\def\adl@drawiv#1#2#3{%
        \hskip.5\tabcolsep
        \xleaders#3{#2.5\@tempdimb #1{1}#2.5\@tempdimb}%
                #2\z@ plus1fil minus1fil\relax
        \hskip.5\tabcolsep}
\newcommand{\cdashlinelr}[1]{%
  \noalign{\vskip\aboverulesep
           \global\let\@dashdrawstore\adl@draw
           \global\let\adl@draw\adl@drawiv}
  \cdashline{#1}
  \noalign{\global\let\adl@draw\@dashdrawstore
           \vskip\belowrulesep}}
\makeatother

\title{Learning Fine-Grained Grounded Citations for \\ Attributed Large Language Models}

\author{
    Lei Huang$^{1}$,
    Xiaocheng Feng$^{1,2}$\thanks{Corresponding Author},
    Weitao Ma$^{1}$, 
    Yuxuan Gu$^{1}$,
    Weihong Zhong$^{1}$,
    Xiachong Feng$^{1}$ \\
    \textbf{Weijiang Yu$^{3}$,
    Weihua Peng$^{3}$,
    Duyu Tang$^{3}$,
    Dandan Tu$^{3}$,
    Bing Qin$^{1,2}$ } \\
    $^{1}$Harbin Institute of Technology, Harbin, China\\
    $^{2}$ Peng Cheng Laboratory\quad \quad $^{3}$Huawei Inc., Shenzhen, China \\
    \texttt{\{lhuang,xcfeng,wtma,yxgu,whzhong,xiachongfeng,qinb\}@ir.hit.edu.cn} \\
    \texttt{\{weijiangyu8,pengwh.hit\}@gmail.com,\{tangduyu,tudandan\}@huawei.com}
}

\begin{document}
\maketitle

\begin{abstract}
Despite the impressive performance on information-seeking tasks, large language models (LLMs) still struggle with hallucinations.
Attributed LLMs, which augment generated text with in-line citations, have shown potential in mitigating hallucinations and improving verifiability.
However, current approaches suffer from suboptimal citation quality due to their reliance on in-context learning.
Furthermore, the practice of citing only coarse document identifiers makes it challenging for users to perform fine-grained verification.
In this work, we introduce \textbf{\textsc{Front}}, a training framework designed to teach LLMs to generate \textbf{F}ine-g\textbf{R}ained gr\textbf{O}u\textbf{N}ded ci\textbf{T}ations.
By grounding model outputs in fine-grained supporting quotes, these quotes guide the generation of grounded and consistent responses, not only improving citation quality but also facilitating fine-grained verification.
Experiments on the ALCE benchmark demonstrate the efficacy of \ours in generating superior grounded responses and highly supportive citations. With LLaMA-2-7B, the framework significantly outperforms all the baselines, achieving an average of 14.21\% improvement in citation quality across all datasets, even surpassing ChatGPT\footnote{Our data and code can be found at: \url{https://github.com/LuckyyySTA/Fine-grained-Attribution}.}.
\end{abstract}
\section{Introduction}
The recent advent of large language models (LLMs) \citep{touvron2023llama2, openai2023gpt4, zhao2023a} has taken the world by storm, fueling a paradigm shift in information acquisition \citep{zhu2023large}.
Despite their compelling performance, LLMs still struggle with hallucinations \citep{ji2023survey, huang2023survey}, a tendency to fabricate non-existent facts or generate unfaithful content.
This issue further poses a risk of misinformation dissemination \citep{chen2023combating}, directly impacting the reliability and trustworthiness of LLMs. 

\begin{figure}[t]
    \centering
    \includegraphics[width=0.5\textwidth]{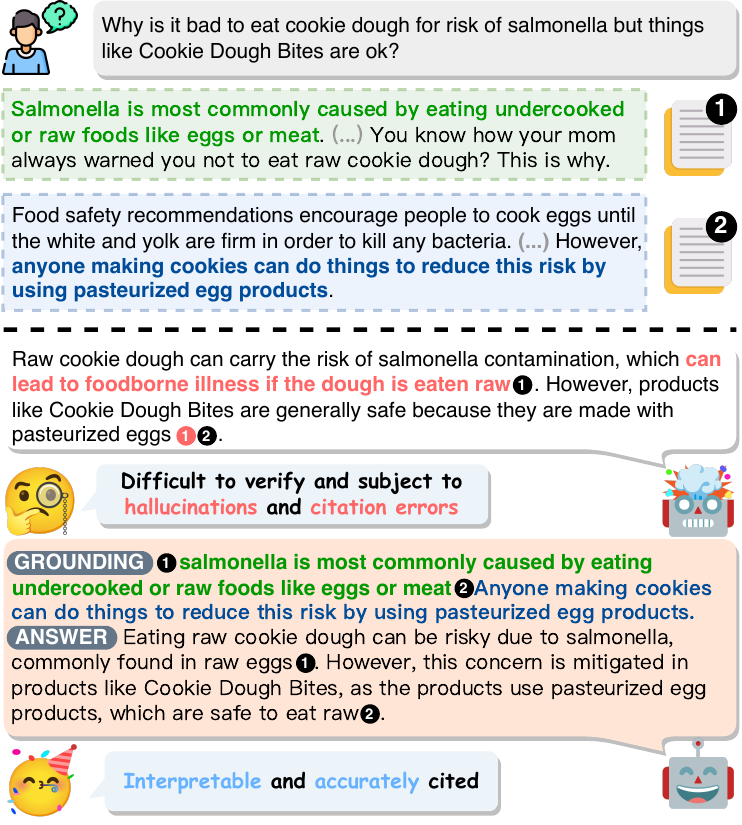}
    \vspace{-0.3in}
    \caption{Compared with the current attributed systems, the core idea behind \ours is to first select the supporting quotes from retrieved sources and then condition the generation process on them, ensuring grounded responses and accurate citations.}
    \vspace{-0.1in}
    \label{fig:example}
\end{figure}

Such prevalence of hallucinations in LLM outputs has motivated the development of attributed systems \citep{nakano2021webgpt, thoppilan2022lamda, menick2022teaching}, such as New Bing\footnote{\url{https://www.bing.com/chat}} and Perplexity\footnote{\url{https://www.perplexity.ai}}, where LLMs are allowed to generate responses with in-line citations. 
Not only does it improve factuality and alleviate hallucinations, but it also simplifies user verification of model outputs, further enhancing the verifiability of LLMs.

Despite recent advancements, current attributed LLMs still expose significant limitations. 
\textbf{Firstly}, current approaches predominantly rely on either in-context learning \citep{gao2023alce} or post-hoc retrieval \citep{gao2023rarr} to achieve attribution, lacking an inherent attribution capability within LLMs, thereby resulting in compromised citation quality \citep{liu2023evaluating}.
\textbf{Secondly}, these citations are typically presented in the form of either document identifiers \citep{nakano2021webgpt} or URLs \citep{thoppilan2022lamda}. Such coarse attribution complicates the process for users to perform fine-grained verification (\eg, pinpoint specific supporting evidence), particularly in lengthy documents. 

To this end, we aim to advance attributed text generation by empowering LLMs with fine-grained attribution ability.
However, one challenge comes from the acquisition of high-quality attribution data for supervised fine-tuning, which is difficult and costly to annotate, and therefore scarce.
Thus, we start with an automatic data generation pipeline tailored for collecting high-quality attribution data (\S\ref{ssec: data}).
Given a user query, the pipeline automates data construction through document retrieval, relevance reranking, attributed answer generation, and data filtering to ensure both the informativeness and attributability of the answers.
Furthermore, to better unlock LLMs' ability for fine-grained attribution, we introduce \ours, a two-stage training framework that teaches LLMs to generate \textbf{F}ine-g\textbf{R}ained gr\textbf{O}u\textbf{N}ded ci\textbf{T}ations (\S\ref{ssec: training}), consisting of Grounding Guided Generation (\bm{$G^3$}) and Consistency-Aware Alignment (\textbf{CAA}).
During the \bm{$G^3$} stage, the LLM first selects supporting quotes from retrieved sources (\textit{grounding}) and then conditions the generation process on them (\textit{generation}).
The \textbf{CAA} stage then utilizes preference optimization to further align the \textit{grounding} and \textit{generation} process by automatically constructing preference signals.
In this way, these quotes can serve as fine-grained citations and improve the efficiency of the verification process for users (see Figure~\ref{fig:example}).

We conduct extensive experiments to evaluate our framework on the ALCE Benchmark \citep{gao2023alce}. Our findings are as follows: 
\begin{itemize}
	\item \ours demonstrates supervisor performance gains in citation quality compared to all baselines, achieving an average 14.21\% improvement using LLaMA-2-7B.
	\item Human evaluation reveals that the quotes generated by our framework are of high quality and significantly benefit user verification.
	\item Analysis shows that \ours generates less hallucination and demonstrates remarkable generalization across different base models.
\end{itemize}   
\section{Related Work}

\paragraph{Retrieval Augmented Generation.}
Recently, retrieval augmented generation (RAG) \citep{karpukhin2020dense, lewis2020retrieval, feng2023trends, gao2023retrieval} has shown promise in knowledge-intensive tasks.
By incorporating retrieved documents, LLMs are equipped with up-to-date information, significantly mitigating knowledge gaps. 
However, recent studies \citep{shi2023large, yoran2023making, xu2023recomp, zhu2024information} have revealed that existing retrieval-augmented LLMs struggle to handle irrelevant or contradictory retrieval documents and effectively utilize contextual evidence.
These limitations can result in performance degradation or even hallucinations \citep{huang2023survey}, highlighting the necessity for more factual and verifiable systems.

\paragraph{Attributed Large Language Models.}
The persistent challenge of hallucinations within LLMs has spurred the development of attributed LLMs \citep{bohnet2022attributed, li2023a, worledge2023unifying}, which seek to enhance information verifiability by generating responses with attribution to evidence sources.
The way of providing attributions varies across studies. 
For example, \citet{gao2023alce} enables LLMs to generate text with in-line citations via in-context learning. 
Another line of research \citep{gao2023rarr, xu2023search} explores post-hoc attribution, where LLMs first generate an initial response and then retrieve the most relevant evidence to achieve attribution.
In this paper, we advance the research on attributed LLMs further. Unlike existing models that predominantly cite document identifiers, we delve into a more fine-grained form of attribution by pinpointing and citing specific extractive quotes.
\section{Task Formulation and Methodology}
\begin{figure*}[h]
    \centering
    \includegraphics[width=1.0\textwidth]{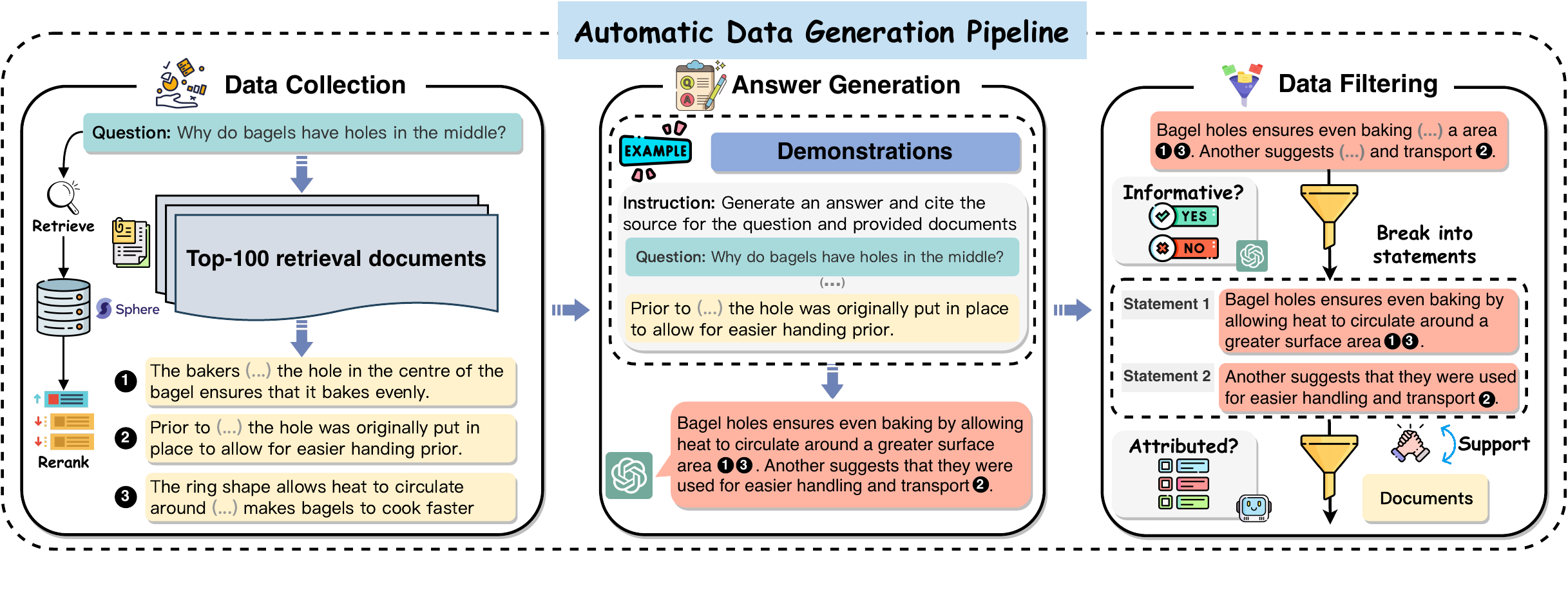}
    \vspace{-0.4in}
    \caption{Overview of the data generation pipeline. The pipeline consists of three primary steps: data collection, answer generation, and data filtering. Firstly, given a user query, the data collection module retrieves the top 100 relevant documents and employs a reranking model to select the top 5 most pertinent documents. Subsequently, attributed responses are generated by distilling ChatGPT via in-context learning. Finally, all responses are filtered by the data filtering module to ensure informativeness and attributability.}
    \vspace{-0.1in}
    \label{fig:pipeline}
\end{figure*}

Following \citep{liu2023evaluating, gao2023alce}, the task is formalized as follows: given a user query $q$ and a corpus of retrieved documents $\mathcal{D}$ as input, the LLM is required to produce a response $\mathcal{S}$, which consists of statements with embedded in-line citations.
We assume the response $\mathcal{S}$ comprising with $n$ statements $\mathcal{S} = \{s_1, s_2, \ldots, s_n \}$ and each statement $s_i \in \mathcal{S}$, cites a list of passage $\mathcal{C}_i = \{c_{i1}, c_{i2}, \ldots\}$, where $c_{ij} \in \mathcal{D}$.
Specifically, citations are presented in the form of \texttt{[1][2]}.

Next, we present a comprehensive overview of our method, which consists of two primary components: an automatic data generation pipeline (\S\ref{ssec: data}) and a two-stage training framework (\S\ref{ssec: training}).

\subsection{Automatic Data Generation Pipeline}
\label{ssec: data}
Equipping LLMs with the attribution capability necessitates training data that includes high-quality responses paired with precise citations, which is typically labor-intensive and costly.
To address this challenge, we propose a pipeline designed for the automatic generation of high-quality attributed data\footnote{\textit{Attributed data} refers to ``answers with in-line citations''.}. This pipeline comprises three core components: data collection, attributed answer generation, and data filtering, as outlined in Figure~\ref{fig:pipeline}.

\paragraph{Data Collection.}
To simulate the real-world environment for information-seeking, we collect questions from the AQuAMuSe dataset \citep{kulkarni2020aquamuse}, which is derived from the Natural Question (NQ) dataset \citep{kwiatkowski2019natural}. The NQ dataset comprises real user queries from the Google search engine, providing a robust basis for realistic question-answering scenarios.
The dataset spans a range of diverse question types, demanding answers of varying lengths, from concise to detailed.
To mimic the way a search engine might synthesize documents of high relevance in response to a user query, we employ Sphere \citep{piktus2021the}, a pre-processed and cleaned version of the Common Crawl corpus, serving as a proxy web search index.
In particular, for a given user query sampled from the AQuAMuSe dataset, we initially retrieve the top 100 relevant documents from the Sphere corpus using sparse retrieval.
These documents are subsequently re-ranked by RankVicuna \citep{pradeep2023rankvicuna} considering its superior performance in listwise re-ranking, resulting in the top 5 most relevant documents for each query.

\paragraph{Attributed Answer Generation.}
Given the remarkable performance of ChatGPT in attributed question answering, we employ ChatGPT to generate answers with corresponding citations for given queries and the top 5 retrieved documents. We provide precise instructions and in-context demonstrations to ensure that ChatGPT produces informative responses and cites the sources accordingly.

\begin{figure*}[h]
    \centering
    \includegraphics[width=1.0\textwidth]{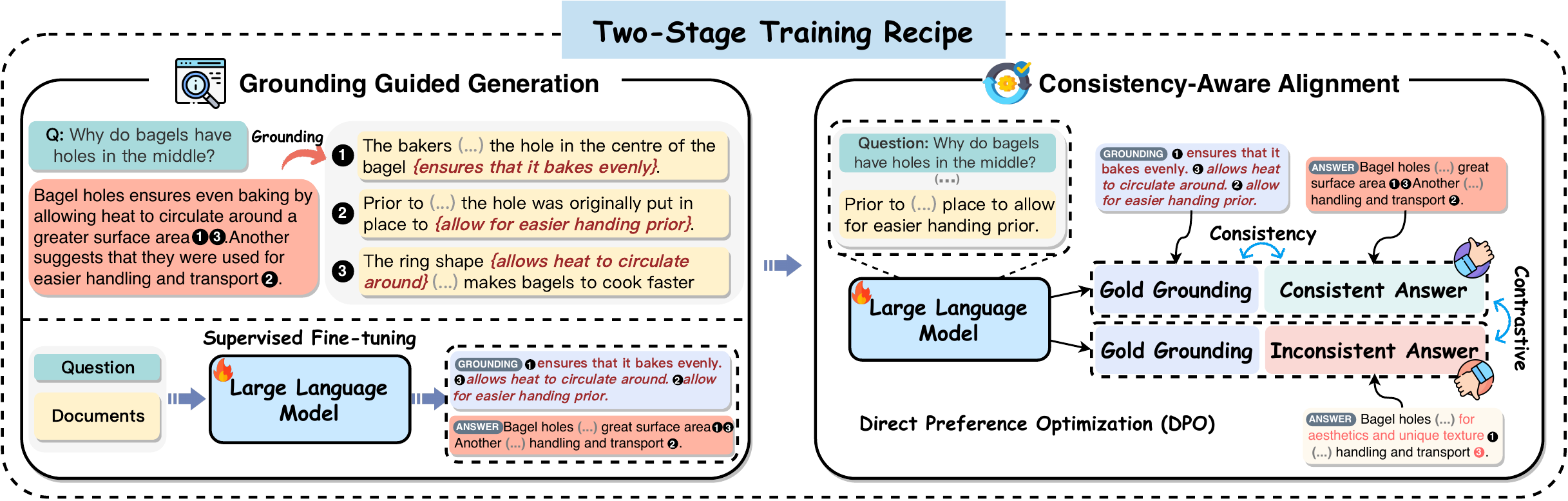}
    \caption{Overview of \ours: The training recipe consists of two stages: grounding-guided generation and consistency-aware alignment. It enables LLMs to first generate precise grounding and subsequently guide the generation of attributed answers, thereby enhancing fine-grained attribution capability.}
    \label{fig:framework}
\end{figure*}

\paragraph{Data Filtering.}
To guarantee the high quality of our synthetic training data, we employ a data filtering process guided by two key criteria derived from \citet{kamalloo2023hagrid}: (1) \textit{informativeness}: assessing if the answer provides sufficient information to the question, and (2) \textit{attributability}: determining if the answer is attributed to the cited documents. 
To mitigate the impact of nonsensical queries and irrelevant document retrieval that may lead to non-informative answers, we utilize ChatGPT for preliminary informativeness annotations.
Responses categorized as non-informative are directly excluded.
Furthermore, to ensure that answers are accompanied by highly supportive citations, we train a discriminator on human-labeled data from the comprehensive evaluation by \citet{liu2023evaluating}, where attributability is categorized into three levels: full support, partial support, or no support.
We quantitatively map the discriminator's outputs to an attributability score and ultimately derive an average score for each attributed answer. Answers falling below a defined threshold are systematically excluded to ensure the synthetic data's reliability, which results in nearly 8,000 entries.
For more details, please refer to Appendix \ref{appendix:data_generation}.

\subsection{Two-Stage Training Recipe}
\label{ssec: training}
In this section, we introduce \ours, a two-stage training framework that aims at empowering LLMs with fine-grained attribution capability.
Figure~\ref{fig:framework} illustrates the overview of our framework.

\subsubsection{Grounding Guided Generation}
To empower LLMs with fine-grained attribution capability, we propose \textbf{G}rounding \textbf{G}uided \textbf{G}eneration (\textbf{$\bm G^3$}), which teaches LLMs to generate fine-grained citations. The cornerstone of \textbf{$\bm G^3$} lies in enabling LLMs to extract supporting quotes from the source documents, each associated with its document identifier, which in turn guides the generation of attributed answers.
Such a format offers two primary benefits. Firstly, the direct extraction of quotes from sources significantly reduces the impact of the incorporation of irrelevant information and the risk of hallucinations in subsequent attributed answers. Secondly, the process naturally facilitates accurate attribution, with each document identifier serving as a clear supervised signal that delineates the origin of the extractive quotes, thus improving the citation quality. 

However, the absence of specific grounding content for statements within our generated dataset poses additional challenges. To tackle this, we employ ChatGPT to meticulously extract segments from cited documents that support the corresponding statement.
Hence, when given a query $q$ and the top-5 retrieved documents $\mathcal{D}$ as input, the LLM is fine-tuned to generate a response $\mathcal{S}$ which consists of two components: the grounded quotes $\mathcal{G}$ and the attributed answer $\mathcal{A}$. 
Specifically, the grounded quotes $\mathcal{G}$ are delineated as follows:
\begin{equation}
	\begin{aligned}
    \mathcal{G} = \{\texttt{[GROUNDING]}, (i_1, e_1), \ldots, (i_n, e_n)\},
    \label{eqn:grounding}
    \end{aligned}
\end{equation}
where \texttt{[GROUNDING]} denotes a special token indicating the start of the grounding process. Each tuple within $\mathcal{G}$, comprising a document identifier $i$ and the corresponding extractive segment $e$, collectively forming a grounded quote.

Similarly, the formulation of the attributed answer $\mathcal{A}$ is concisely presented as:
\begin{equation}
	\begin{aligned}
    \mathcal{A} = \{\texttt{[ANSWER]}, s_1, s_2 \ldots, s_m\},
    \label{eqn:answer}
    \end{aligned}
\end{equation}
where \texttt{[ANSWER]} is a special token that signals the beginning of the answer generation process. Each statement $s_i$ cites a list of passages $\mathcal{C}_i = \{c_{i1}, c_{i2}, \ldots\}$, where $c_{ij} \subseteq \{i_1, i_2, \ldots, i_n\}$, as defined in Equation \ref{eqn:answer}.

Thus, the training loss is formulated as:
\begin{equation}
    \mathcal{L} = -\sum_{i=1}^{N} \log P(y_i | q_i, \mathcal{D}_i; \theta)
\end{equation}
where $y_i$ represents the combined output of grounded quotes $\mathcal{G}$ and the answer $\mathcal{A}$ for each given query $q_i$ and set of retrieved documents $\mathcal{D}_i$.
\subsubsection{Consistency-Aware Alignment}
While $\bm{G^3}$ unlocks the ability to first extract supporting quotes before generating attributed answers, it occasionally leads to inconsistencies between grounded quotes and attributed answers. Such discrepancies challenge the attempt to employ these grounded quotes as fine-grained verification. In response, we propose a consistency-aware alignment (\textbf{CCA}) stage specifically aimed at enhancing the consistency between the grounding process and the generation process.

The cornerstone of our approach involves contrasting a consistent answer with an inconsistent one under the guidance of the same oracle grounded quotes. This aligns with the concept of Reinforcement Learning from Human Feedback (RLHF) \citep{ouyang2022training, bai2022training}, where LLMs are further fine-tuned to distinguish between desirable and undesirable responses under preference feedback. However, such contrastive preference feedback typically comes from human annotation. 
To automatically construct preference pairs for preference optimization, we utilize the attributed answers generated by smaller LLMs (\eg, LLaMA-2-7B) under the in-context learning setting as negative samples.
These answers, characterized by their low quality and inconsistency with oracle grounded quotes, automatically serve as contrastive supervision signals when paired with high-quality attributed answers labeled in \S\ref{ssec: data}. 
In this scenario, the process not only encourages the LLM to generate attributed answers that are more consistent with the grounded quotes but also facilitates the identification and correction of nuanced errors present in smaller models.

Specifically, we adopt Direct Preference Optimization \citep{rafailov2023direct}, a variant of RLHF known for its stability, for our contrastive alignment.
Formally, for each instance, given the oracle grounded $g^{(i)}$ along with a consistent oracle answer $y_w^{(i)}$ as well as an attributed answer $y_l^{(i)}$ generated by a weaker LLM via in-context learning, we can simply construct a preference dataset:
\begin{equation}
    \mathcal{D}=\bigl\{x^{(i)}, \tau_w^{(i)}, \tau_l^{(i)}\bigr\}_{i=1}^N,
    \label{eqn:data}
\end{equation}
where $\tau_w^{(i)} = g^{(i)} \circ y_w^{(i)}$ denotes the concatenation of the oracle grounding with the consistent, attributed answer, $\tau_l^{(i)} = g^{(i)} \circ y_l^{(i)}$ denotes the concatenation with the inconsistent attributed answer. Here, $\circ$ signifies the operation of string concatenation.

Finally, we can optimize the policy model $\pi_\theta$ on the dataset $\mathcal{D}$ by minimizing the following loss: 
\begin{equation}
    \begin{aligned}
        &\mathcal{L}_{\mathrm{DPO}}(\pi_\theta; \pi_{ref}; \mathcal{D}) \\
        =&  -\mathbb{E}_{(x,\tau_w,\tau_l)\sim \mathcal{D}} \bigg[ \log \sigma \bigg( \beta \log \frac{\pi_\theta(\tau_w | x)}{\pi_{\mathrm{ref}}(\tau_w | x)} \\
& - \beta \log \frac{\pi_\theta(\tau_l | x)}{\pi_{\mathrm{ref}}(\tau_l | x)} \bigg) \bigg],
    \label{eqn:loss}
    \end{aligned}
\end{equation}
where $\pi_\mathrm{ref}$ represents the reference model, initialized from $\bm{G^3}$. The hyper-parameter $\beta$ modulates the divergence between the distribution from the policy model and the reference model. $\tau_w$ is the consistent answer, while $\tau_l$ is the inconsistent one.

\section{Experimental Settings}

\subsection{Datasets}
We conduct experiments on the ALCE benchmark \citep{gao2023alce}, designed for attributed text generation.
The benchmark includes three long-form QA datasets that span various types of questions. 

\paragraph{ASQA} \citep{stelmakh2022asqa} is a long-form factoid QA dataset characterized by inherently ambiguous questions that require multiple short answers to encapsulate different viewpoints.
\paragraph{ELI5} \citep{fan2019eli5} features open-ended questions intended for simplification to the comprehension level of five-year-olds, requiring explanatory multi-sentence responses.
\paragraph{QAMPARI} \citep{amouyal2022qampari} is a factoid QA dataset derived from Wikipedia, where answers are structured as a compilation of entities.
\definecolor{color3}{HTML}{FEFAE0}
\newcolumntype{a}{>{\columncolor{green!10}}c}
\newcolumntype{b}{>{\columncolor{blue!10}}c}
\newcolumntype{d}{>{\columncolor{color3}}c}

\begin{table*}[t]
\small
  \centering
  \resizebox{\linewidth}{!}{
    \begin{tabular}{llaaaabbbbddddd}
    \toprule
    \multirow{2}[4]{*}{\textbf{Model Type}} & \multirow{2}[4]{*}{\textbf{Model Size}} & \multicolumn{4}{c}{\textbf{ASQA}} & \multicolumn{4}{c}{\textbf{ELI5}} & \multicolumn{4}{c}{\textbf{QAMPARI}}\\
    \cmidrule(lr){3-6}
    \cmidrule(lr){7-10}
    \cmidrule(lr){11-15}
    & &  \multicolumn{1}{c}{\textbf{Correctness}} & \multicolumn{3}{c}{\textbf{Citation}}  & \multicolumn{1}{c}{\textbf{Correctness}} & \multicolumn{3}{c}{\textbf{Citation}} &  \multicolumn{2}{c}{\textbf{Correctness}} &  \multicolumn{3}{c}{\textbf{Citation}} \\
    \cmidrule(lr){3-3}
    \cmidrule(lr){4-6}
    \cmidrule(lr){7-7}
    \cmidrule(lr){8-10}
    \cmidrule(lr){11-12}
    \cmidrule(lr){13-15}
    & & \multicolumn{1}{c}{EM Rec.} & \multicolumn{1}{c}{Rec.} & \multicolumn{1}{c}{Prec.} & \multicolumn{1}{c}{F1.} & \multicolumn{1}{c}{Claim} & \multicolumn{1}{c}{Rec.} & \multicolumn{1}{c}{Prec.} & \multicolumn{1}{c}{F1} & \multicolumn{1}{c}{Rec.-5} & \multicolumn{1}{c}{Prec.} & \multicolumn{1}{c}{Rec.} & \multicolumn{1}{c}{Prec.} & \multicolumn{1}{c}{F1} \\
    \midrule
    \headercolor
    \multicolumn{15}{c}{\textbf{\textit{Prompting-based}}} \\
    \midrule
    ChatGPT & - & 40.37 & 72.81 & 69.69 & 71.22 & 12.47 & 49.44 & 47.05 & 48.22 & 20.28 & 19.84 & 19.06 & 22.03 & 20.44 \\
    \cdashlinelr{1-15}
    \multirow{3}{*}{LLaMA-2} & 7B & 24.32 & 17.24 & 17.87 & 17.55 & 4.53 & 3.92 & 5.38 & 4.54 & 12.56 & 11.32 & 6.03 & 6.35 & 6.19 \\
    & 13B & 27.99 & 16.45 & 19.04 & 17.65 & 7.77 & 8.49 & 8.43 & 8.46 & 18.00 & 12.39 & 5.45 & 5.74 & 5.59 \\
    & 70B & 31.53 & 44.18 & 44.79 & 44.48 & 10.43 & 23.75 & 22.43 & 23.07 & 18.50 & 14.79 & 10.10 & 10.50 & 10.30 \\
    \cdashlinelr{1-15}
    \multirow{3}{*}{LLaMA-2-Chat} & 7B & 29.93 & 55.99 & 51.66 & 53.74 & 12.47 & 19.90 & 15.48 & 17.41 & 17.96 & 19.74 & 9.58 & 9.68 & 9.63 \\
    & 13B & 34.39 & 37.15 & 38.17 & 37.65 & 13.83 & 16.50 & 16.09 & 16.29 & 21.34 & 18.86 & 8.94 & 9.06 & 9.00 \\
    & 70B & 41.24 & 60.19 & 61.16 & 60.67 & 13.30 & 36.63 & 36.63 & 36.63 & 22.62 & 18.04 & 13.49 & 13.98 & 13.73 \\
    \cdashlinelr{1-15}
    \multirow{2}{*}{Vicuna-v1.5} & 7B & 38.34 & 48.37 & 44.63 & 46.42 & 12.30 & 29.81 & 22.45 & 25.61 & 14.22 & 14.74 & 11.26 & 11.64 & 11.45 \\
    & 13B & 35.20 & 51.92 & 53.40 & 52.65 & 14.33 & 31.15 & 28.99 & 30.03 & \ul{22.06} & 19.60 & 13.04 & 13.74 & 13.38 \\
    \cdashlinelr{1-15}
    \multirow{2}{*}{Mistral} & 7B & 29.46 & 23.12 & 25.45 & 24.23 & 8.47 & 16.04 & 16.32 & 16.18 & 16.96 & 15.98 & 7.50 & 7.76 & 7.63 \\
    & 8 $\times$ 7B & 36.30 & 32.72 & 34.49 & 33.58 & 10.43 & 26.11 & 25.09 & 25.59 & 18.18 & 15.63 & 9.72 & 10.20 & 9.95 \\
    \cdashlinelr{1-15}
    \multirow{2}{*}{Mistral-Instruct} & 7B & 38.57 & 64.90 & 59.67 & 62.18 & 11.07 & 49.25 & 42.69 & 45.74 & 17.52 & 21.29 & 17.56 & 18.53 & 18.03 \\
    & 8 $\times$ 7B & \textbf{44.11} & 61.80 & 63.27 & 62.53 & 13.93 & 49.28 & 48.34 & 48.81 & 20.12 & 19.64 & 19.27 & 20.38 & 19.81 \\
    \midrule
     \headercolor
    \multicolumn{15}{c}{\textbf{\textit{Post-hoc Retrieval}}} \\
    \midrule
        ChatGPT & - & 37.68 & 27.11 & 27.05 & 27.08 & \textbf{18.77} & 14.55 & 14.55 & 14.55 & \textbf{25.14} & \textbf{22.85} & 12.29 & 12.29 & 12.29 \\
        \cdashlinelr{1-15}
        LLaMA-2-Chat & 70B & 29.68 & 24.51 & 24.51 & 24.51 & 16.03 & 12.93 & 12.93 & 12.93 & 17.90 & 14.45 & 9.05 & 9.05 & 9.05 \\
        \cdashlinelr{1-15}
        Mistral-Instruct & 8 $\times$ 7B & 33.90 & 24.57 & 24.48 & 24.52 & \ul{17.37} & 15.68 & 15.68 & 15.68 & \ul{24.16} & 18.28 & 9.78 & 9.78 & 9.78 \\
    \midrule
    \headercolor
    \multicolumn{15}{c}{\textbf{\textit{Training-based}}} \\
    \midrule
    \multirow{2}{*}{Self-RAG (LLaMA-2)} & 7B & 29.96 & 67.82 & 66.97 & 67.39 & 6.90 & 22.34 & 32.40 & 26.45 & 2.34 & 1.98 & 10.53 & 18.80 & 13.50 \\
    & 13B & 31.66 & 71.26 & 70.35 & 70.80 & 6.07 & 30.46 & 40.20 & 34.66 & 1.90 & 1.33 & 12.79 & 20.90 & 15.86 \\
    \cdashlinelr{1-15}
    \multirow{2}{*}{\vani-SFT (LLaMA-2)} & 7B & 40.32 & 67.67 & 63.67 & 65.61 & 9.63 & 42.30 & 40.06 & 41.15 & 12.86 & 21.09 & 21.35 & 21.36 & 21.35 \\
    & 13B & 40.85 & 71.49 & 66.21 & 68.75 & 10.27 & 46.75 & 44.47 & 45.58 & 12.68 & \ul{22.80} & 23.64 & 23.71 & 23.67 \\
    \cdashlinelr{1-15}
    \multirow{2}{*}{\ours (LLaMA-2)} & 7B & 40.84 & \ul{77.70} & \ul{69.89} & \ul{73.59} & 9.18 & \ul{58.60} & \ul{55.33} & \ul{56.92} & 11.50 & 21.38 & \ul{24.74} & \ul{24.84} & \ul{24.79} \\
    & 13B & \ul{41.51} & \textbf{78.44} & \textbf{73.66} & \textbf{75.97} & 9.32 & \textbf{60.31} & \textbf{59.21} & \textbf{59.75} & 11.94 & 22.61 & \textbf{24.86} & \textbf{25.39} & \textbf{25.12} \\
    \bottomrule
    \end{tabular}%
  }
  \caption{Main results on the ALCE benchmark. \textbf{Bold} numbers indicate the best performance, while $\_$ indicates the second-best performance.}
  \label{tab:alce_result}
\end{table*}%

\subsection{Evaluation Metrics}
Following the ALCE benchmark \citep{gao2023alce}, our evaluation primarily focuses on two key dimensions: \textbf{Citation Quality} and \textbf{Correctness}. Detailed descriptions of additional evaluation dimensions are presented in the Appendix \ref{appendix:alce_evaluation}.
\paragraph{Citation Quality.} Citation quality is critical for evaluating LLM attribution, assessed along two dimensions: (1) \textit{Citation Recall}, determining if the output is entirely supported by the cited documents, and (2) \textit{Citation Precision}, assessing if each citation supports its corresponding statement. Evaluation is conducted by TRUE \citep{honovich2022true}, a T5-11B model fine-tuned on a collection of NLI datasets to automatically examine the entailment of cited documents and the model generation. Additionally, to capture a holistic measure of citation quality, we also report the \textit{Citation F1}, the harmonic mean of citation precision and recall:
\begin{equation}
	\begin{aligned}
    F_1 = 2 \cdot \frac{\textrm{citation precision} \cdot \textrm{citation recall}}{\textrm{citation precision} + \textrm{citation recall}},
    \label{eqn:citation_f1}
    \end{aligned}
\end{equation}
\paragraph{Correctness.} 
For the ASQA dataset, correctness is quantified using exact match recall (\textbf{EM Rec.}) by checking whether the short answers are exact substrings of the generation.
Regarding the ELI5 dataset, correctness is measured through claim recall (\textbf{Claim}), evaluating whether the model's response entails the ground truth sub-claims. 
For the QAMPARI dataset, correctness is assessed using exact match precision (\textbf{Prec.}) and top-5 exact match recall (\textbf{Rec.-5}) \textemdash \ considered 100\% if the prediction includes at least five correct answers.
\subsection{Baselines}
We compare our method with three types of baselines: prompting-based, post-hoc retrieval, and training-based.

\subsubsection{Prompting-based Methods.}
We directly prompt LLMs using few-shot demonstrations, each consisting of a query, the top 5 relevant retrieved documents, and an answer with in-line citations.
Our experiments encompass a range of LLMs, from foundational models to supervised fine-tuning (SFT) LLMs. 
For foundational LLMs, we select \textbf{GPT-3.5-Turbo}\footnote{Specifically, we utilize \texttt{gpt-3.5-turbo-1106} version} as the representative closed-source model, recognized for its notable performance.
Among the open-source foundational LLMs, we focus on the LLaMA-2 series including \textbf{LLaMA2-7B}, \textbf{LLaMA2-13B}, and \textbf{LLaMA2-70B}, as well as the Mistral series, which spans from \textbf{Mistral-7B} \citep{jiang2023mistral} to \textbf{Mistral-8x7B-MoE} \citep{jiang2024mixtral}. 
Regarding SFT LLMs, we select the SFT counterparts of the open-source foundational LLMs we used.
Detailed prompting settings can be found in Appendix \ref{appendix:appendix_prompt}. 

\subsubsection{Post-hoc Retrieval Methods.}
Following \citet{gao2023alce}, we first instruct LLMs to answer the given query in a closed-book setting, and then integrate citations in a post-hoc manner.
For each generated statement, we employ GTR \citep{ni2022large} to identify and cite the most relevant document from the top 100 retrieved documents.
We utilize the same models mentioned in prompting-based settings for this baseline.

\subsubsection{Training-based Methods.}
\paragraph{Self-RAG} \citep{asai2023selfrag} Self-RAG trains the LLM to learn to adaptively retrieve passages on-demand and enable it to reflect on its generation to further improve generation quality and attributions.
\paragraph{\vani-SFT} We directly employ supervised fine-tuning to train the LLM on our generated training data. Given a query and corresponding documents, the LLM is required to directly generate answers with citations.

\subsection{Implement Details}
We implement \ours with different sizes of foundational models (LLaMA-2-7B and LLaMA-2-13B) to evaluate its effectiveness.
During the evaluation, \ours utilize the same retrieval settings as those outlined by \citet{gao2023alce}. Additional details of training and evaluation settings can be found in Appendix \ref{appendix:appendix_exp}.

\section{Results and Analysis}
\subsection{Overall Results}
\label{ssec:overall}

\definecolor{color3}{HTML}{FEFAE0}
\newcolumntype{a}{>{\columncolor{green!10}}c}
\newcolumntype{b}{>{\columncolor{blue!10}}c}
\newcolumntype{d}{>{\columncolor{color3}}c}

\begin{table*}[t]
\small
  \centering
  \resizebox{\linewidth}{!}{
    \begin{tabular}{laaaabbbbddddd}
    \toprule
    \multirow{2}[4]{*}{\textbf{Model}} & \multicolumn{4}{c}{\textbf{ASQA}} & \multicolumn{4}{c}{\textbf{ELI5}} & \multicolumn{4}{c}{\textbf{QAMPARI}}\\
    \cmidrule(lr){2-5}
    \cmidrule(lr){6-9}
    \cmidrule(lr){10-14}
    &  \multicolumn{1}{c}{\textbf{Correctness}} & \multicolumn{3}{c}{\textbf{Citation}}  & \multicolumn{1}{c}{\textbf{Correctness}} & \multicolumn{3}{c}{\textbf{Citation}} &  \multicolumn{2}{c}{\textbf{Correctness}} &  \multicolumn{3}{c}{\textbf{Citation}} \\
    \cmidrule(lr){2-2}
    \cmidrule(lr){3-5}
    \cmidrule(lr){6-6}
    \cmidrule(lr){7-9}
    \cmidrule(lr){10-11}
    \cmidrule(lr){12-14}
    & \multicolumn{1}{c}{EM Rec.} & \multicolumn{1}{c}{Rec.} & \multicolumn{1}{c}{Prec.} & \multicolumn{1}{c}{F1.} & \multicolumn{1}{c}{Claim} & \multicolumn{1}{c}{Rec.} & \multicolumn{1}{c}{Prec.} & \multicolumn{1}{c}{F1} & \multicolumn{1}{c}{Rec.-5} & \multicolumn{1}{c}{Prec.} & \multicolumn{1}{c}{Rec.} & \multicolumn{1}{c}{Prec.} & \multicolumn{1}{c}{F1} \\
    \midrule
     \ours-7B & 40.84 & 77.70 & 69.89 & 73.59 & 9.18 & 58.60 & 55.33 & 56.92 & 11.50 & 21.38 & 24.74 & 24.84 & 24.79 \\
     \textsc{Self-Guide} (w/o Consistency) & 38.99 & 70.69 & 64.48 & 67.44 & 10.04 & 47.63 & 44.80 & 46.17& 12.18 & 20.03 & 22.50 & 22.58 & 22.54 \\
     \vani-SFT (w/o Ground) & 40.32 & 67.67 & 63.67 & 65.61 & 9.63 & 42.30 & 40.06 & 41.15 & 12.86 & 21.09 & 21.35 & 21.36 & 21.35 \\
    \midrule
    \ours-13B & 41.51 & 78.44 & 73.66 & 75.97 & 9.32 & 60.31 & 59.21 & 59.75 & 11.94 & 22.61 & 24.86 & 25.39 & 25.12 \\
    \textsc{Self-Guide} (w/o Consistency) & 40.99 & 73.08 & 68.13 & 70.52 & 10.06 & 50.68 & 49.78 & 50.23 & 13.94 & 22.38 & 23.73 & 23.99 & 23.85 \\
     \vani-SFT (w/o Ground) & 40.85 & 71.49 & 66.21 & 68.75 & 10.27 & 46.75 & 44.47 & 45.58 & 12.68 & 22.80 & 23.64 & 23.71 & 23.67 \\
    \bottomrule
    \end{tabular}%
  }
  \caption{Ablation study on the impact of different training stages within the ALCE benchmark.}
  \label{tab:ablation}
\end{table*}%

\paragraph{Simply supervised fine-tuning can boost citation quality.}
As shown in Table \ref{tab:alce_result}, teaching LLMs to generate responses with citations via supervised fine-tuning significantly enhances citation quality, demonstrating substantial improvements over both prompt-based and post-hoc retrieval baselines across all datasets. 
Specifically, with LLaMA-2-7B, \vani-SFT led to substantial gains in citation F1 scores over prompting: ASQA (17.55 $\rightarrow$ 65.61), ELI5 (4.54 $\rightarrow$ 41.15), and QAMPARI (6.19 $\rightarrow$ 21.35).
These gains highlight the effectiveness of our training data generation pipeline.

\paragraph{\ours achieves significant performance gains and surpasses ChatGPT.}
While \vani-SFT demonstrates strong performance, it still shows notable discrepancies compared to leading open-source LLMs, such as Mixtral-8×7B-Instruct (\eg, 41.15 vs. 45.74) and ChatGPT (\eg, 41.15 vs. 48.22) on the ELI5 dataset. \ours not only bridges these gaps but also establishes significant leads across all datasets. Specifically, using LLaMA-2-7B, \ours comprehensively outperforms ChatGPT, achieving increases of 3.32\%, 18.04\%, and 21.28\% in citation quality on the ASQA, ELI5, and QAMPARI datasets respectively. This performance underscores the effectiveness of \ours in enhancing attribution capabilities.

\paragraph{\ours exhibits scalability with model size.}
As illustrated at the bottom of Table \ref{tab:alce_result}, the performance of \ours in terms of citation quality shows notable improvements when scaling from 7B to 13B. Specifically, we observe improvements of 3.23\% in ASQA, 4.97\% in ELI5, and 1.33\% in QAMPARI. This upward trend underscores the scalability of \ours with increasing model size, demonstrating the potential of \ours in leveraging the increased capabilities of larger LLMs for further performance gains.

\paragraph{\ours demonstrates remarkable generalization.}
Compared to the varied queries and answer types present in the ALCE benchmark, our training data, derived exclusively from the AQuAMuSe dataset \citep{kulkarni2020aquamuse}, exhibits out-of-domain characteristics. Nonetheless, \ours demonstrates superior citation quality, affirming its exceptional ability to generalize across diverse query types and retrieval documents. Additionally, while not specifically optimized for correctness, \ours also showcases modest improvements in this metric over \vani-SFT on the ASQA and QAMPARI datasets. However, \ours encounters lower Rec.-5 on the QAMPARI dataset, likely due to the nature of its answers, which consist of concatenated entities, diverging significantly from our training data.

\subsection{Ablation Study}
We conduct ablation studies to verify the effectiveness of different components proposed in \ours. 

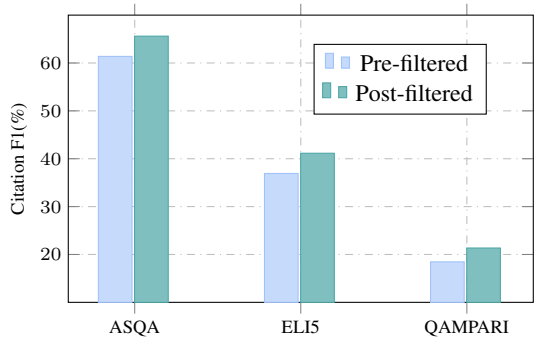
\begin{figure}
    \centering
    \begin{tikzpicture}
    \normalsize{
    \scriptsize{
      \begin{axis}[
      at={(0em,0em)},
    ymajorgrids=true,
        xmajorgrids=true,
        grid style=dashdotted,
        ybar,
        enlarge x limits=0.2,
        height=.7\linewidth,
        width=1.0\linewidth,
        bar width=1.8em,
        symbolic x coords = {{ASQA},{ELI5},{QAMPARI}},
        ymin=10,
        ymax=70,
        ytick = {20,30,40,50,60},
        ylabel={Citation F1(\%)},
        xtick = data,
        xticklabel style={/pgf/number format/fixed,/pgf/number format/fixed zerofill,/pgf/number format/precision=1},
        legend style={at={(0.9,0.9)}, anchor=north east, font=\small}
        ]
        \addplot[fill=blue!30, draw=blue!50] coordinates {({ASQA},61.38) ({ELI5},36.92) ({QAMPARI},18.46)};
        \addplot[fill=teal!50, draw=teal!70, xshift=-0.15em] coordinates {({ASQA},65.61) ({ELI5},41.15) ({QAMPARI},21.35)};

  \addlegendentry{Pre-filtered}
  \addlegendentry{Post-filtered}

      \end{axis}
    }
    }
    \end{tikzpicture}
    \caption{Ablation Study on Data Filtering.}
    \label{fig:ablation_data_filtering}
    \end{figure}

\paragraph{Effects of Data Generation Pipeline.}
As illustrated in \S\ref{ssec:overall}, simply SFT achieves strong performance, underscoring the high quality of our training data. Furthermore, data filtering, a crucial component of our data generation pipeline, plays a pivotal role in ensuring the quality of the generated data by filtering out queries that yield non-informative answers or fail to meet attribution criteria. To validate the effectiveness of our data filtering strategies, we conducted experiments comparing models fine-tuned on both pre-filtered and post-filtered data. The results, depicted in Figure \ref{fig:ablation_data_filtering}, confirm that models trained on filtered data exhibit a notable improvement in citation quality over those trained on unfiltered data, achieving superior attribution performance with reduced data volume.

\paragraph{Effects of Grounding Guided Generation ($\bm{G^3}$).}
\definecolor{c2}{HTML}{378BA4}
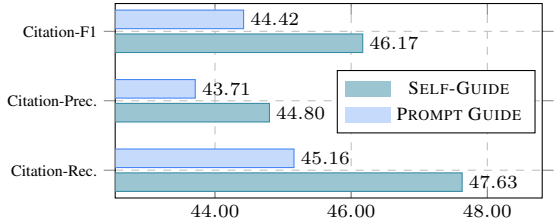
\begin{figure}[!t]
\centering
\begin{tikzpicture}
    \tiny{
    \scriptsize{
    \begin{axis}[
      at={(0em,0em)},
      ymajorgrids,
      xmajorgrids,
      grid style=dashed,
      xbar,
      height=.26\textwidth,
      width=.45\textwidth,
      bar width=1em,
      legend image code/.code={%
                    \draw[#1, draw=none] (0cm,-0.1cm) rectangle (0.6cm,0.1cm);
                },
      legend style={at={(18.2em,5.4em)}},
      symbolic y coords={{Citation-Rec.}, {Citation-Prec.}, {Citation-F1}},
      yticklabel style={align=right,font=\tiny},
      ytick=data,
      nodes near coords,
      nodes near coords style={/pgf/number format/.cd,fixed zerofill,precision=2},
      nodes near coords align={horizontal},
      enlarge y limits=0.2,yticklabel style={/pgf/number format/fixed,/pgf/number format/fixed zerofill,/pgf/number format/precision=2},
      enlarge x limits=0.3,xticklabel style={/pgf/number format/fixed,/pgf/number format/fixed zerofill,/pgf/number format/precision=2},]
      \addplot[fill=c2!50, draw=c2] coordinates {(47.63,{Citation-Rec.}) (44.80,{Citation-Prec.}) (46.17,{Citation-F1})};
      \addplot[fill=blue!30, draw=blue] coordinates {(45.16,{Citation-Rec.}) (43.71,{Citation-Prec.}) (44.42,{Citation-F1})};
      \addlegendentry{\textsc{Self-Guide}}
      \addlegendentry{\textsc{Prompt Guide}}

    \end{axis}
    }
  }
\end{tikzpicture}
\caption{Ablation study of different grounding guidance strategies on the ELI5 dataset.}
    \label{fig:ablation_grounding}
\vspace{-1.1em}
\end{figure}

$\bm{G^3}$ empowers LLMs to first select relevant fine-grained quotes, which subsequently guide the generation process. These quotes can provide fine-grained supervision signals for attributed text generation.
To evaluate the effectiveness of $\bm{G^3}$, we conduct an ablation study comparing it against two variants with distinct training recipes.
Given that \ours consists of two stages, we refer to the model trained only during the first stage (without consistency-aware alignment) as \textsc{Self-Guide}.
We first compare \textsc{Self-Guide} against \vani-SFT (w/o Ground), which is trained to directly generate responses with citations, bypassing the grounding step. The ablation study, detailed in Table \ref{tab:ablation}, reveals that models incorporating grounding guidance significantly outperform their \vani-SFT counterparts that lack such grounding mechanisms. This highlights the crucial role of grounding in bolstering attribution.

Moreover, we explore an alternative variant of grounding guidance.
Considering that \textsc{Self-Guide} leverages the model itself to both select grounded quotes and generate attributed answers in an end-to-end paradigm, a natural variant involves breaking down this task into two distinct stages. In this variant, ChatGPT is tasked with extracting grounded quotes. Subsequently, a separate model is trained to utilize these grounded quotes, along with the query and retrieval documents, to directly output the response and citations. This variant, referred to as \textsc{Prompt-Guided}, integrates grounded quotes into the prompt to guide the generation process.
Experiments conducted on the ELI5 dataset using the LLaMA-2-7B model show that \textsc{Self-Guide} outperforms \textsc{Prompt-Guide}. Results depicted in Figure \ref{fig:ablation_grounding} indicate that training models to self-generate grounded quotes before generating attributed responses is more effective than simply incorporating these grounded quotes into the prompt.
\definecolor{bg_color}{RGB}{234,234,242}
\definecolor{color1}{HTML}{457b9d}
\definecolor{color2}{HTML}{F6AA1C}
\definecolor{color3}{HTML}{FDF0D5}
\definecolor{color4}{HTML}{A8DADC}
\definecolor{color5}{HTML}{C6AC8F}
\definecolor{color6}{HTML}{B8BEDD}

\begin{figure}[!t]
\centering
\begin{tikzpicture}
\scriptsize{
  \begin{axis}[
      width=.30\textwidth,
      height=.24\textwidth,
      xtick={20,30,40,50,60},
      xmin=10, xmax=65,
      scale only axis,
      xticklabel style={rotate=45,anchor=east,align=center, font=\scriptsize},
      ymin=1.25,
      ymax=3.25,
      xlabel={Citation F1},
      ylabel={Faithfulness},
      extra y ticks={1.5,2,2.5,3}, 
      extra x ticks={20,30,40,50,60}, 
      extra tick style={grid=major},
      grid style={dashed,gray!50},
      axis background/.style={fill=bg_color},
   ]
  
  \addplot+[only marks, mark=*, color=color1, mark options={scale=1}] coordinates {(17.41, 1.64)};
  \node at (axis cs:17.41,1.64) [anchor=west] {\tiny{LLaMA2-7B-Chat}};

  \addplot+[only marks, mark=*, color=color2, mark options={scale=1}] coordinates {(56.92, 2.77)};
  \node at (axis cs:56.92,2.75) [anchor=east] {\tiny{\ours-7B}};
  
  \addplot+[only marks, mark=*, color=color2, mark options={scale=1}] coordinates {(46.17, 2.28)};
  \node at (axis cs:46.17,2.32) [anchor=east] {\tiny{\textsc{Self-Guide}-7B}};
  
  \addplot+[only marks, mark=*, color=color3, mark options={scale=2}] coordinates {(30.03, 2.20)};
  \node at (axis cs:30.03,2.15) [anchor=north] {\tiny{Vicuna-13B}};
  
  \addplot+[only marks, mark=*, color=color4, mark options={scale=2}] coordinates {(59.75, 2.99)};
  \node at (axis cs:57.2,2.99) [anchor=east] {\tiny{\ours-13B}};
  
    \addplot+[only marks, mark=*, color=color4, mark options={solid, scale=2}] coordinates {(50.23, 2.49)};
  \node at (axis cs:50.23,2.53) [anchor=south] {\tiny{\textsc{Self-Guide}-13B}};
  
  \addplot+[only marks, mark=*, color=color5, mark options={solid, scale=3}] coordinates {(48.81, 2.36)};
  \node at (axis cs:48.81,2.26) [anchor=north] {\tiny{Mixtral-8x7B-Inst}};
  
  \addplot+[only marks, mark=*, color=color6, mark options={solid, scale=4}] coordinates {(36.63, 2.57)};
  \node at (axis cs:34.58,2.57) [anchor=east] {\tiny{LLaMA2-70B-Chat}};
  \end{axis}
 }
\end{tikzpicture}
\caption{The relationship between citation F1 and hallucination: Models positioned closer to the top-right corner exhibit higher citation quality and a lower degree of hallucination.}
\label{fig:trade-off}
\end{figure}
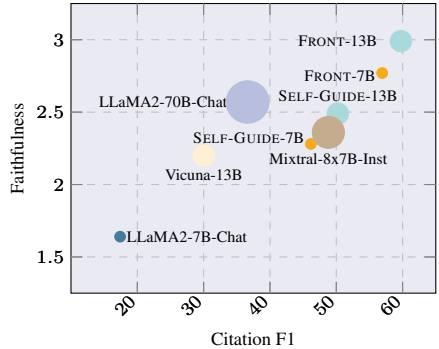

\paragraph{Effects of Consistency-Aware Alignment (CCA).}
The primary goal of \textbf{CCA} is to enhance the consistency between grounded quotes and attributed answers, thereby alleviating hallucinations and achieving more precise attribution. To evaluate this, we compare models that underwent only the $\bm{G^3}$ stage (\textsc{Self-Guide}) with those further enhanced through the CCA stage (\ours). As illustrated in Table \ref{tab:ablation}, \ours significantly improves citation quality over \textsc{Self-Guide}, demonstrating the effectiveness of the CCA stage in enhancing attribution. 

Furthermore, to assess \textbf{CCA}'s impact on reducing hallucinations, we utilize QAFactEval \citep{fabbri2022qafacteval}, a widely used metric for factual consistency, which scores the consistency of model responses to given documents on a scale from 0 to 5, with higher scores indicating greater faithfulness. Specifically, we analyze the performance of leading open-source models and two variants of \ours and \textsc{Self-Guide} on the ELI5 dataset. As shown in Figure \ref{fig:trade-off}, \ours produces more faithful outputs than \textsc{Self-Guide}, significantly reducing hallucinations. 

\paragraph{Effects of Training Data Scale.}
We analyze the impact of the data scale on model performance across two training stages. In particular, we randomly sampled 2k, 4k, 6k, and 8k instances from our full training data across two distinct training stages.  These subsets were then utilized to fine-tune various 7B model variants, enabling a comparative analysis of performance based on data scale. Results are shown in Figure \ref{fig:mixture_of_data}, which indicates that increasing data size shows significant enhancements in citation quality, indicating a positive correlation between data size and model performance.
As \ours implements an automated procedure capable of generating high-quality attributed data and constructing contrastive supervision from weak and strong LLMs, it holds the potential for continuous performance improvements.
\section{Human Evaluation}
Given the significant impact of the quality of grounded quotes on fine-grained verification for users, we conducted a human evaluation to assess the quality of grounded quotes at different stages of our framework: (1) Quotes extracted by ChatGPT from 50 sampled data points during the $\bm{G^3}$ stage. (2) Quotes generated by \ours-7B across three datasets, with 50 data points sampled from each.

We engaged four annotators, each with relevant expertise and holding at least a bachelor's degree.
The quality of quotes was evaluated on two dimensions: \textbf{authenticity} and \textbf{helpfulness}. Authenticity (a binary scale of 0/1) refers to whether the quotes genuinely originate from the corresponding documents (quotes that are hallucinated or mismatched with the corresponding document ID are considered inauthentic). Helpfulness (5-point Likert scale) refers to the degree to which the quotes are beneficial in addressing the query. The results in Table \ref{tab:human-quotes} represent the average scores for all quotes within each model response, with two annotators evaluating each response to ensure reliability.

Furthermore, to evaluate the consistency of quote quality annotations, we computed the inter-annotator agreement using Fleiss' Kappa coefficient. The obtained Kappa coefficient of 0.82 indicates a high level of agreement among annotators. The results of the human evaluation indicate that both quotes extracted by ChatGPT and those generated by \ours are of high quality, further substantiating the effectiveness of our method.

\begin{table}[t]
    \centering
    \resizebox{0.98\linewidth}{!}{
    \begin{tabular}{lcc}
        \toprule
            & Authenticity & Helpfulness \\
            \midrule
            ChatGPT & 0.93 & 4.08 \\
            \ours-7B on ASQA & 0.94 & 3.86 \\
            \ours-7B on ELI5 & 0.92 & 3.96 \\
            \ours-7B on QAMPARI & 0.86 & 3.62 \\
            \bottomrule
            \end{tabular}
    }
    \caption{
        Human evaluation on the quality of grounded quotes. 
    }
    \vspace{-10pt}
    \label{tab:human-quotes}
\end{table}

\section{Conclusion}
In this work, we present \ours, a two-stage training framework designed to equip LLMs with fine-grained attribution capabilities.
\ours enables LLMs to initially select supporting quotes, which then guide the generation process.
By further enhancing the consistency between the grounding and generation process via preference optimization, these supporting quotes can serve as fine-grained citations.
Through comprehensive experiments, \ours has demonstrated its ability to generate superior grounded responses and highly supportive citations.
Further analysis shows that \ours significantly reduces hallucinations and benefits user verification.

\section{Limitation}
Our study presents several limitations worth noting. Firstly, the validation of our framework is predominantly conducted on models of sizes 7B and 13B, leaving the exploration of larger models, such as LLaMA-2 70B due to computational constraints.
Secondly, our framework relies on a prior retrieval process, wherein relevant documents are retrieved at one time. The incorporation of adaptive retrieval, enabling more dynamic interactions with LLMs, could potentially enhance performance. We leave it for future research.
Lastly, evaluating the correctness of long-form question answering presents inherent challenges, leading our framework to primarily enhance citation quality, with modest advancements in correctness. Therefore, we advocate for the development of more robust metrics capable of accurately assessing the correctness of long-form QA responses, paving the way for future work.

\section*{Acknowledgements}
We appreciate Yuchun Fan for providing valuable suggestions. Xiaocheng Feng is the corresponding author of this work. We thank the anonymous reviewers for their insightful comments. This work was supported by the National Key R\&D Program of China via grant No. 2021ZD0112905, the National Natural Science Foundation of China (NSFC) (grant 62276078, U22B2059), the Key R\&D Program of Heilongjiang via grant 2022ZX01A32, the International Cooperation Project of PCL, PCL2022D01 and the Fundamental Research Funds for the Central Universities (Grant No.HIT.OCEF.2023018).

\bibliography{custom}

\appendix
\section{Details of Data Generation Pipeline}
\label{appendix:data_generation}

\subsection{Data Statistic}
\label{appendix:data_statistic}
\begin{table}[ht]
\centering
\resizebox{.48\textwidth}{!}{ 
\begin{tabular}{lc}
\toprule
\#~Questions & 8,098\\
\cdashlinelr{1-2}
\ding{229} \#~Long Answer & 5667\\
\ding{229} \#~Short Answer & 2431\\
\hline
Avg. Words per Answer & 50.48\\
\cdashlinelr{1-2}
\ding{229} Avg. Words per Long Answer & 69.15\\
\ding{229} Avg. Words per Short Answer & 6.94\\
\midrule
Avg. Citation per Answer & 4.40\\
\cdashlinelr{1-2}
\ding{229} Avg. Citation per Long Answer& 4.68\\
\ding{229} Avg. Citation per Short Answer& 3.77\\
\bottomrule
\end{tabular}
}
\caption{
The statistics of the data generated by our automatic data generation pipeline.
}
\label{table:stats}
\end{table} 
Table \ref{table:stats} presents the statistics of the data automatically generated by our data generation pipeline. In total, we collected 8,098 questions from the Natural Questions (NQ) dataset, of which 5,667 questions were gathered from those with long-form answers, and 2,431 questions were collected from those with short-form factoid answers.

For questions requiring long-form answers, we initialized our query source with the AQUAMUSE dataset \citep{kulkarni2020aquamuse}, which consists of high-quality queries specifically designed for long-form responses within the NQ dataset, recognized as “good” by the majority of NQ evaluators. In this way, utilizing a refined and superior quality query set laid a robust groundwork for our training data generation, streamlining the data filtering process. For factoid queries that necessitate short-form answers, we directly sampled from the original NQ dataset, leveraging its abundance and inherently high quality.

During the data generation process, our initial query set comprised 7,725 queries requiring long-form answers and 4,000 queries necessitating short-form answers. After a two-stage data filtering process, we retained 5,667 and 2,431 queries, respectively. Additionally, we calculated the average length of answers and the average number of citations generated for various types of queries within our dataset, as shown in Table \ref{table:stats}.

\subsection{Details of Data Filtering}
\label{appendix:detail_attributed_dis}
We trained our Attributed Discriminator using the manually annotated data provided by \citet{liu2023evaluating}, which is sampled from real generative search engines. Each statement and its cited document have been meticulously annotated for attribution, categorized into three types: complete support, partial support, and no support. For training, we utilized a dataset of 8,834 instances, comprising 6,415 instances of complete support, 1,552 of partial support, and 867 of no support. The discriminator initialized with LLaMA-2-7B, was trained with a maximum sequence length of 512. We trained it for 3 epochs, with a total batch size of 128, and a peak learning rate of 2e-5, incorporating 3\% warmup steps, followed by a linear decay.

During the data filtering stage, we first break down the automatically generated attributed answers into statement form and use the trained discriminator to annotate the attribution between each statement and its cited documents. Specifically, we assign different attribution scores to each statement $s$ based on its attribution relationship with cited documents $d$, as shown in Equation \ref{equ:reward_function}. Consequently, for each attributed answer, we can calculate its average attribution score. Attributed answers with an average attribution score below 0.8 are filtered out. The threshold of 0.8 was determined through preliminary testing on the development set, for which we manually annotated 100 samples to ensure the effectiveness of our filtering criteria.

\begin{equation}
\begin{aligned}
    r(s)\! =\! 
        \begin{cases} 
            1, &  \!\!\!\!\! \texttt{Dis}(s,d) = \texttt{complete support} \\
            0.5, &\!\!\!\!\! \texttt{Dis}(s,d) = \texttt{partial support} \\
            0, & \!\!\!\!\! \texttt{Dis}(s,d) = \texttt{no support}
    \end{cases}
    \label{equ:reward_function}
 \end{aligned}
\end{equation}


\section{Details of Evaluation Metrics}
\label{appendix:alce_evaluation}
In addition to evaluating citation quality and correctness, the ALCE benchmark includes a broader set of dimensions, such as fluency, ROUGE-L, and generation length.
\paragraph{Fluency}
We evaluate the fluency of the generated response using MAUVE \citep{pillutla2021mauve}. Notably, we calculate fluency only for the ASQA and ELI5 datasets, omitting it for QAMPARI, as the response in QAMPARI typically consists of lists of short answers. A relatively high MAUVE score indicates that the generation is sufficiently fluent.
\paragraph{ROUGE-L}
In addition to evaluating the correctness of the model-generated content, we employ ROUGE-L to assess the overall quality and textual coherence of the responses.

\section{Prompts}
\label{appendix:appendix_prompt}

\subsection{Prompts for Prompting-based Methods}
Following \citet{gao2023alce}, we adopt the vanilla prompting strategy for its simplicity and effectiveness. Specifically, the prompts vary according to the type of data within the ALCE benchmark. For long-form QA datasets such as ASQA and ELI5, the prompt format is detailed in Table \ref{tab:prompt_long}. For the short-form QA dataset QAMPARI, the format is outlined in Table \ref{tab:prompt_short}.

\begin{table*}[t]
    \centering
    \small
    \begin{tabular}{>{\raggedright\arraybackslash\tt}p{0.95\textwidth}<{}}
        \toprule
            Instruction: Write an accurate, engaging, and concise answer for the given question using only the provided search results (some of which might be irrelevant) and cite them properly. Use an unbiased and journalistic tone. Always cite for any factual claim. When citing several search results, use [1][2][3]. Cite at least one document and at most three documents in each sentence. If multiple documents support the sentence, only cite a minimum sufficient subset of the documents. \\
        \bottomrule
    \end{tabular}
    \caption{
        Prompt for Long-form QA.
    }
    \label{tab:prompt_long}
\end{table*}

\begin{table*}[t]
    \centering
    \small
    \begin{tabular}{>{\raggedright\arraybackslash\tt}p{0.95\textwidth}<{}}
        \toprule
            Instruction: Provide a list of accurate answers for the given question using only the provided search results (some of which might be irrelevant) and cite them properly. Always cite one and only one document for each answer. Separate answers by commas. For questions that have more than 5 answers, write at least 5 answers. \\
        \bottomrule
    \end{tabular}
    \caption{
        Prompt for Short-form QA.
    }
    \label{tab:prompt_short}
\end{table*}

\subsection{Instructions for \ours}

\begin{table*}[t]
    \centering
    \small
    \begin{tabular}{>{\raggedright\arraybackslash\tt}p{0.95\textwidth}<{}}
        \toprule
        	Below is an instruction that describes a task, paired with an input that provides further context. Write a response that appropriately completes the request.
        	
        	\\
        	\#\#\# Instruction: \\
        	
            Extract the relevant content from the provided documents and then use the extracted content to guide answer generation and cite the sources properly. \\
            \#\#\# Input:Question: \{Question\} Documents: \{Documents\}
            
            \#\#\# Response: \\
        \bottomrule
    \end{tabular}
    \caption{
        Instruction Format for \ours on Long-form QA.
    }
    \label{tab:instruct_long}
\end{table*}

\begin{table*}[t]
    \centering
    \small
    \begin{tabular}{>{\raggedright\arraybackslash\tt}p{0.95\textwidth}<{}}
        \toprule
        	Below is an instruction that describes a task, paired with an input that provides further context. Write a response that appropriately completes the request.
        	
        	\\
        	\#\#\# Instruction: \\
        	
            Extract the relevant content from the provided documents and then use the extracted content to provide a list of accurate answers for the given question. Always cite one and only one document for each answer. Separate answers by commas. \\
            \#\#\# Input:Question: \{Question\} Documents: \{Documents\}
            
            \#\#\# Response: \\
        \bottomrule
    \end{tabular}
    \caption{
        Instruction Format for \ours on Short-form QA.
    }
    \label{tab:instruct_short}
\end{table*}

During the training process, we follow the instruction format of Alpaca\footnote{\url{https://github.com/tatsu-lab/stanford_alpaca/tree/main}}. Specifically, we employ varied instructions for different question types, as delineated in Table \ref{tab:instruct_long} for long-form questions and Table \ref{tab:instruct_short} for short-form questions.

\section{Experimental Details}
\label{appendix:appendix_exp}

\subsection{Training Details of \ours}
\label{appendix:detail_training}
The training of all models is executed on 4 Nvidia A100 GPUs, each with 80GB of memory, leveraging the Deepspeed \citep{rasley2020deepspeed} and HuggingFace Accelerate libraries \citep{accelerate} to conduct multi-GPU distributed training. Given the long nature of the inputs, the maximum token length is set to 2,048 tokens.

During the grounding guide generation stage, models are trained for 5 epochs with a total batch size of 128, a peak learning rate of 2e-5 with 3\% warmup steps followed by a linear decay. 
During the contrastive alignment stage, we set the $\beta$ to 0.1 and continued training for two additional epochs. Specifically, 
During inference, we use the vllm framework \citep{kwon2023efficient} for efficient inference. The hyperparameters are set as illustrated in Table \ref{appendix:params}.

\subsection{Retrieval Settings}
\label{appendix:detail_retrieval}
During the evaluation, we adopt the same retrieval settings as specified by \citet{gao2023alce}. For the ASQA and QAMPARI datasets, we use the dense retriever GTR \citep{ni2022large}. For the ELI5 dataset, we employ the sparse retriever BM25.

\begin{table}[t]
\centering
\scalebox{0.8}{
\begin{tabular}{l|r}
\toprule
\textbf{Hyper-parameters}
& \textbf{Value}
\\
\midrule
Top-p  & 0.95 \\
Temperature & 0.2	\\
Max-length & 2048 \\
\bottomrule
\end{tabular}
}
\small
\caption{
Hyper-parameter settings in inference.
}
\label{appendix:params}
\end{table}

\section{More detail about Ablation Study}

\subsection{The Effect of Training Data Scale.}
\definecolor{tiffanyblue}{RGB}{129,216,208}
\definecolor{bangdiblue}{RGB}{0,149,182}
\definecolor{kleinblue}{RGB}{0,47,167}
\definecolor{kabuliblue}{RGB}{26,85,153}
\definecolor{purple}{RGB}{138,43,226}

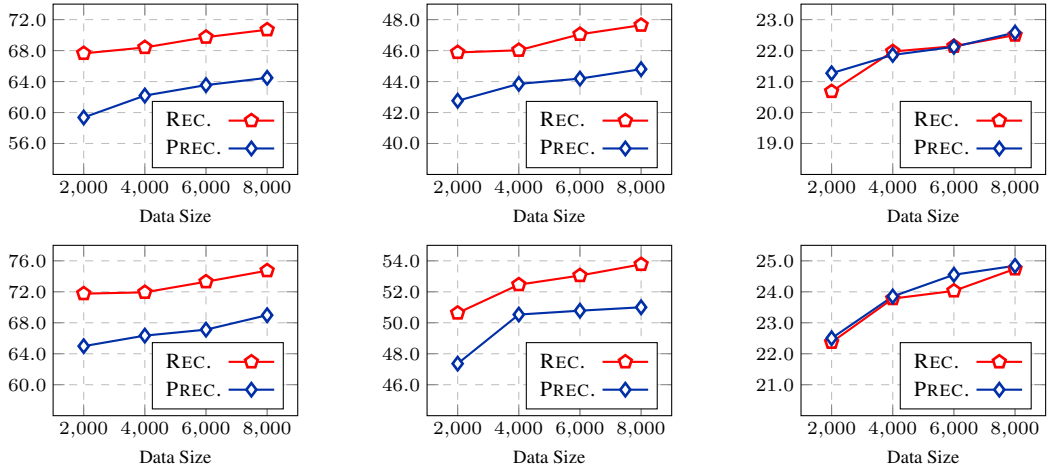
\begin{figure*}[!t]
  \centering
  \begin{tikzpicture}[]
    \scriptsize{
      \begin{axis}[
	 at={(0em,13em)},
      ymajorgrids,
      xmajorgrids,
      grid style=dashed,
      width=.30\textwidth,
      height=.24\textwidth,
      legend style={at={(0.48,0.13)}, anchor=south west},
      xlabel={\scriptsize{Data Size}},
      ylabel={\scriptsize{}},
      ylabel style={yshift=-2em},xlabel style={yshift=0.0em},
      yticklabel style={/pgf/number format/precision=1,/pgf/number format/fixed zerofill},
      ymin=52,ymax=74, ytick={56,60,64,68,72},
      xmin=1000,xmax=9000,xtick={2000,4000,6000,8000},
      legend style={yshift=-6pt,xshift=-1em, legend plot pos=right,font={\footnotesize},cells={anchor=west}}
      ]

      \addplot[red,mark=pentagon*,,mark size=2.5pt,thick,mark options={fill=white,draw=red,line width=1.0pt}] coordinates {(2000,67.64) (4000,68.40) (6000,69.73) (8000,70.69)};
      \addlegendentry{\scalebox{.9}{\textsc{Rec.}}}

      \addplot[kleinblue,mark=diamond*,mark size=2.5pt,thick,mark options={fill=white,draw=kleinblue,line width=1.0pt}] coordinates {(2000,59.36) (4000,62.19) (6000,63.54) (8000,64.48)
      };
      \addlegendentry{\scalebox{.9}{\textsc{Prec.}}}

      \end{axis}
     }
	\scriptsize{
      \begin{axis}[
	 at={(20em,13em)},
      ymajorgrids,
      xmajorgrids,
      grid style=dashed,
      width=.30\textwidth,
      height=.24\textwidth,
      legend style={at={(0.48,0.13)}, anchor=south west},
      xlabel={\scriptsize{Data Size}},
      ylabel={\scriptsize{}},
      ylabel style={yshift=-2em},xlabel style={yshift=0.0em},
      yticklabel style={/pgf/number format/precision=1,/pgf/number format/fixed zerofill},
      ymin=38,ymax=49, ytick={40,42,44,46,48},
      xmin=1000,xmax=9000,xtick={2000,4000,6000,8000},
      legend style={yshift=-6pt,xshift=-1em, legend plot pos=right,font={\footnotesize},cells={anchor=west}}
      ]
      \addplot[red,mark=pentagon*,mark size=2.5pt,thick,mark options={fill=white,draw=red,line width=1.0pt}] coordinates { (2000,45.88)  (4000,46.02) (6000,47.05) (8000,47.64)
      };
      \addlegendentry{\scalebox{.9}{\textsc{Rec.}}}

      \addplot[kleinblue,mark=diamond*,mark size=2.5pt,thick,mark options={fill=white,draw=kleinblue,line width=1.0pt}] coordinates {(2000, 42.76) (4000,43.85) (6000,44.19) (8000,44.80)};
      \addlegendentry{\scalebox{.9}{\textsc{Prec.}}}
      \end{axis}
     }
     
  	\scriptsize{
      \begin{axis}[
	 at={(40em,13em)},
      ymajorgrids,
      xmajorgrids,
      grid style=dashed,
      width=.30\textwidth,
      height=.24\textwidth,
      legend style={at={(0.48,0.13)}, anchor=south west},
      xlabel={\scriptsize{Data Size}},
      ylabel={\scriptsize{}},
      ylabel style={yshift=-2em},xlabel style={yshift=0.0em},
      yticklabel style={/pgf/number format/precision=1,/pgf/number format/fixed zerofill},
      ymin=18,ymax=23.5, ytick={19,20,21,22,23},
      xmin=1000,xmax=9000,xtick={2000,4000,6000,8000},
      legend style={yshift=-6pt,xshift=-1em, legend plot pos=right,font={\footnotesize},cells={anchor=west}}
      ]

      \addplot[red,mark=pentagon*,,mark size=2.5pt,thick,mark options={fill=white,draw=red,line width=1.0pt}] coordinates {(2000,20.68) (4000,21.97) (6000,22.14) (8000,22.50)};
      \addlegendentry{\scalebox{.9}{\textsc{Rec.}}}

      \addplot[kleinblue,mark=diamond*,mark size=2.5pt,thick,mark options={fill=white,draw=kleinblue,line width=1.0pt}] coordinates {(2000,21.27) (4000,21.86) (6000,22.12) (8000,22.58)
      };
      \addlegendentry{\scalebox{.9}{\textsc{Prec.}}}

      \end{axis}
     }
	\scriptsize{
      \begin{axis}[
	 at={(0em,0)},
      ymajorgrids,
      xmajorgrids,
      grid style=dashed,
      width=.30\textwidth,
      height=.24\textwidth,
      legend style={at={(0.48,0.13)}, anchor=south west},
      xlabel={\scriptsize{Data Size}},
      ylabel={\scriptsize{}},
      ylabel style={yshift=-2em},xlabel style={yshift=0.0em},
      yticklabel style={/pgf/number format/precision=1,/pgf/number format/fixed zerofill},
      ymin=56,ymax=78, ytick={60,64,68,72,76},
      xmin=1000,xmax=9000,xtick={2000,4000,6000,8000},
      legend style={yshift=-6pt,xshift=-1em, legend plot pos=right,font={\footnotesize},cells={anchor=west}}
      ]

      \addplot[red,mark=pentagon*,,mark size=2.5pt,thick,mark options={fill=white,draw=red,line width=1.0pt}] coordinates {(2000,71.76) (4000,71.93) (6000,73.3) (8000,74.73)};
      \addlegendentry{\scalebox{.9}{\textsc{Rec.}}}

      \addplot[kleinblue,mark=diamond*,mark size=2.5pt,thick,mark options={fill=white,draw=kleinblue,line width=1.0pt}] coordinates {(2000,64.98) (4000,66.34) (6000,67.11) (8000,68.98)
      };
      \addlegendentry{\scalebox{.9}{\textsc{Prec.}}}

      \end{axis}
     }
	\scriptsize{
      \begin{axis}[
	 at={(20em,0)},
      ymajorgrids,
      xmajorgrids,
      grid style=dashed,
      width=.30\textwidth,
      height=.24\textwidth,
      legend style={at={(0.48,0.13)}, anchor=south west},
      xlabel={\scriptsize{Data Size}},
      ylabel={\scriptsize{}},
      ylabel style={yshift=-2em},xlabel style={yshift=0.0em},
      yticklabel style={/pgf/number format/precision=1,/pgf/number format/fixed zerofill},
      ymin=44,ymax=55, ytick={46,48,50,52,54},
      xmin=1000,xmax=9000,xtick={2000,4000,6000,8000},
      legend style={yshift=-6pt,xshift=-1em, legend plot pos=right,font={\footnotesize},cells={anchor=west}}
      ]
      \addplot[red,mark=pentagon*,mark size=2.5pt,thick,mark options={fill=white,draw=red,line width=1.0pt}] coordinates { (2000,50.63)  (4000,52.46) (6000,53.05) (8000,53.77)
      };
      \addlegendentry{\scalebox{.9}{\textsc{Rec.}}}

      \addplot[kleinblue,mark=diamond*,mark size=2.5pt,thick,mark options={fill=white,draw=kleinblue,line width=1.0pt}] coordinates {(2000, 47.36) (4000,50.53) (6000,50.78) (8000,51.00)};
      \addlegendentry{\scalebox{.9}{\textsc{Prec.}}}
      \end{axis}
     }
     
  	\scriptsize{
      \begin{axis}[
	 at={(40em,0)},
      ymajorgrids,
      xmajorgrids,
      grid style=dashed,
      width=.30\textwidth,
      height=.24\textwidth,
      legend style={at={(0.48,0.13)}, anchor=south west},
      xlabel={\scriptsize{Data Size}},
      ylabel={\scriptsize{}},
      ylabel style={yshift=-2em},xlabel style={yshift=0.0em},
      yticklabel style={/pgf/number format/precision=1,/pgf/number format/fixed zerofill},
      ymin=20,ymax=25.5, ytick={21,22,23,24,25},
      xmin=1000,xmax=9000,xtick={2000,4000,6000,8000},
      legend style={yshift=-6pt,xshift=-1em, legend plot pos=right,font={\footnotesize},cells={anchor=west}}
      ]

      \addplot[red,mark=pentagon*,,mark size=2.5pt,thick,mark options={fill=white,draw=red,line width=1.0pt}] coordinates {(2000,22.37) (4000,23.78) (6000,24.03) (8000,24.74)};
      \addlegendentry{\scalebox{.9}{\textsc{Rec.}}}

      \addplot[kleinblue,mark=diamond*,mark size=2.5pt,thick,mark options={fill=white,draw=kleinblue,line width=1.0pt}] coordinates {(2000,22.50) (4000,23.85) (6000,24.55) (8000,24.84)
      };
      \addlegendentry{\scalebox{.9}{\textsc{Prec.}}}

      \end{axis}
     }

  \end{tikzpicture}
  \caption{Ablation study on synthetic training data size: The upper part of the figure corresponds to the Grounding Guided Generation training stage, while the bottom part represents the Weak-to-Strong Contrastive Alignment training stage. From left to right, the results are presented for ASQA, ELI5, and QAMPARI, respectively. \textsc{Rec.} indicates Citation Recall and \textsc{Prec.} denotes Citation Precision. The x-axis represents the quantity of automatically generated data. It is observed that as the volume of automatically generated data increases, there is a consistent improvement in both citation recall and precision across the two training stages.}\label{fig:mixture_of_data}
\end{figure*}
We examine how model performance varies with changes in data scale, as depicted in Figure\ref{fig:mixture_of_data}. The upper part of the figure illustrates the impact of the training data scale on citation quality during the Grounding Guided Generation training stage, with datasets ASQA, ELI5, and QAMPARI represented from left to right. Similarly, the lower part of the figure describes the influence during the Consistency-Aware Alignment training stage.

\subsection{The Generalization Across Model Architectures.}
\definecolor{bananayellow}{rgb}{1.0, 0.88, 0.21}
\definecolor{blanchedalmond}{rgb}{1.0, 0.92, 0.8}
\definecolor{blue1}{HTML}{A8DADC}
\definecolor{skyblue1}{HTML}{669BBC}
\definecolor{skyblue2}{HTML}{9AC9DB}
\definecolor{darkblue1}{HTML}{22577A}
\definecolor{darkblue2}{HTML}{2878B5}

\pgfplotsset{compat=1.11,
        /pgfplots/ybar legend/.style={
        /pgfplots/legend image code/.code={%
        \draw[##1,/tikz/.cd,bar width=2em,yshift=-0.2em,bar shift=0pt]
                plot coordinates {(0cm,0.8em)};},
},
}
\begin{figure}[!t]
  \centering
    \begin{tikzpicture}  
    \scriptsize{
        \begin{axis}  
        [  
            ybar,
            legend cell align={left},
            ymin=0, ymax=70,
            ytick={0,10,20,30,40,50, 60},
            major x tick style = transparent,
            bar width=10pt,
            width=8cm,
            height=5cm,
            enlarge x limits=0.3,
            ylabel={Citation F1(\%)},
            symbolic x coords={ASQA, ELI5, QAMPARI},  
            xtick=data,  
            axis x line*=bottom,
            axis y line*=left,
            legend style={anchor=east,column sep=10pt,font=\scriptsize,draw=none,xshift=0cm,yshift=0.2cm}
            ]  
        \addplot[ybar, fill=skyblue2,  postaction={}] coordinates {
            (ASQA, 24.23) (ELI5, 16.18) (QAMPARI, 7.63) 
        };
        \addplot[ybar, fill=blue1,  postaction={}] coordinates {
            (ASQA, 46.79) (ELI5, 47.06) (QAMPARI, 11.85)
        };
        \addplot[ybar, fill=darkblue2,  postaction={}] coordinates {
            (ASQA, 66.79) (ELI5, 57.06) (QAMPARI, 21.85)
        };  
        \legend{Mistral-7B, \vani-SFT (Mistral-7B), \ours (Mistral-7B)} 
        \end{axis}  
    }
    \end{tikzpicture}  
    \caption{
    Ablation study on model architecture: We substituted the foundation model in \ours with Mistral-7B and compared the experimental results of models under the same foundation model using in-context learning and those directly supervised fine-tuned on our automatically generated data. The experiments demonstrate that by replacing different foundation models, our framework still maintains its generalizability.}
    \label{fig:model_generalization}
\end{figure}
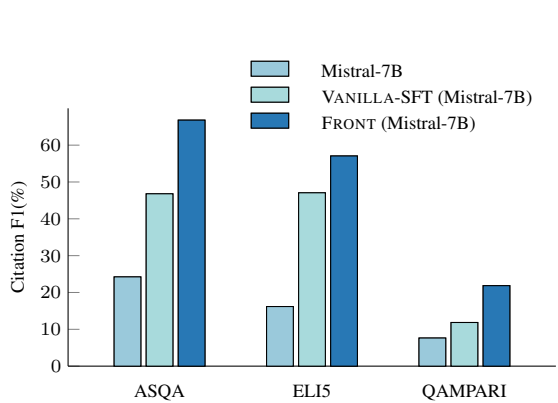

\ours demonstrates exceptional generalization capabilities across various foundational model architectures. Specifically, transitioning the foundational model from LLaMA-2-7B to the stronger foundational model, Mistral-7B, results in even greater performance enhancements as shown in Figure \ref{fig:model_generalization}. This further underscores the broad applicability and generalizability of \ours.

\subsection{The effect of $\beta$ in Consistency-Aware Alignment Training Stage}
In the Consistency-Aware Alignment Training Stage, the $\beta$ parameter in Direct Preference Optimization (DPO) controls the strength of the Kullback-Leibler penalty, typically set within the range of 0.1 to 0.5. A higher $\beta$ value indicates a preference for the policy model's training process to remain closer to the initially referenced model. In extreme cases, as $\beta \rightarrow 0$, we ignore the constraints imposed by the reference model. This setting aims to balance the model's ability to adapt to new training signals while maintaining the stability of the learned behaviors from the reference model.

Subsequently, we trained five variants by adjusting $\beta$ from 0.1 to 0.5 on the model previously trained with $\bm{G^3}$ to explore the impact of the hyperparameter $\beta$ on attribution quality. We evaluated these variants on the ASQA and ELI5 datasets, and the experimental results are shown in Figure \ref{fig:dpo_beta}.

The experimental results indicate that as $\beta$ increases, the model's performance on attribution gradually decreases. This observation suggests that the first stage of $\bm{G^3}$ might introduce a noticeable inconsistency between grounding and attribution. With higher $\beta$ values, the model struggles to escape the constraints of inconsistent attributed answers, leading to a reduction in attribution quality as $\beta$ increases.

\definecolor{Maroon}{HTML}{AE3135}
\definecolor{BLUE}{HTML}{6466AE}
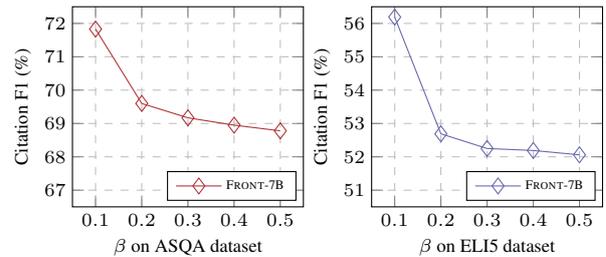
\begin{figure}[t]
    \centering
\pgfplotsset{width=0.6\linewidth,height=0.55\linewidth,compat=1.15}
\footnotesize
\begin{tikzpicture}
\scriptsize{
\begin{axis}[
	at={(0em,0em)},
    xlabel={$\beta$ on ASQA dataset},
    ylabel={Citation F1 (\%)},
    xmin=0.05, xmax=0.55,
    ymin=66.5, ymax=72.5,
    xtick={0.1, 0.2, 0.3, 0.4, 0.5},
    ytick={67, 68, 69, 70, 71, 72},
    legend pos=south east,
    ymajorgrids=true,
    xmajorgrids=true,
    grid style=dashed,
    x label style={at={(axis description cs:0.5,-0.125)},anchor=north},
    y label style={at={(axis description cs:-0.15,0.5)},anchor=south},
    legend style={nodes={scale=0.7, transform shape}}
]
\addplot[
    color=Maroon,
    mark=diamond,
    mark size=3pt,
    ]
    coordinates {
    (0.1, 71.83)
    (0.2, 69.60)
    (0.3, 69.17)
    (0.4, 68.95)
    (0.5, 68.78)
    };
    \addlegendentry{\ours-7B}
\end{axis}
}
\scriptsize{
\begin{axis}[
	at={(16em,0em)},
    xlabel={$\beta$ on ELI5 dataset},
    ylabel={Citation F1 (\%)},
    xmin=0.05, xmax=0.55,
    ymin=50.5, ymax=56.5,
    xtick={0.1, 0.2, 0.3, 0.4, 0.5},
    ytick={51, 52, 53, 54, 55, 56},
    legend pos=south east,
    ymajorgrids=true,
    xmajorgrids=true,
    grid style=dashed,
    x label style={at={(axis description cs:0.5,-0.125)},anchor=north},
    y label style={at={(axis description cs:-0.15,0.5)},anchor=south},
    legend style={nodes={scale=0.7, transform shape}}
]
\addplot[
    color=BLUE,
    mark=diamond,
    mark size=3pt,
    ]
    coordinates {
    (0.1, 56.19)
    (0.2, 52.69)
    (0.3, 52.25)
    (0.4, 52.19)
    (0.5, 52.06)
    };
    \addlegendentry{\ours-7B}
\end{axis}
}
\end{tikzpicture}

    \caption{Ablation on hyperparameter $\beta$ in Weak-to-Strong Contrastive Alignment stage on ASQA and ELI5
    }
    \label{fig:dpo_beta}
\end{figure}

\section{Full Results}
\label{appendix:appendix_result}
We present the comprehensive results of our experiments in Tables \ref{table:asqa_full}, \ref{table:eli5_full}, and \ref{table:qampari_full}. Beyond the evaluation metrics related to Correctness and Citation, we adhere to the evaluation framework established in \citep{gao2023alce}. For long-form QA datasets like ASQA and ELI5, we also report metrics related to Fluency, ROUGE-L, and average response length. Specifically, we use MAUVE \citep{pillutla2021mauve} to evaluate the fluency of the model response. For datasets like QAMPARI, where answers are composed of concatenated entities, we calculate the average number of predicted entities.

\begin{table*}[h]
    \centering
    \small
        \begin{tabular}{lllcccccc}
            \toprule
            \multirow{2}[4]{*}{\textbf{Model Type}} & \multirow{2}[4]{*}{\textbf{Model Size}} & \textbf{Fluency} & \textbf{Correct.} & \multicolumn{3}{c}{\textbf{Citation}}\\
            \cmidrule(lr){3-3} \cmidrule(lr){4-4} \cmidrule(lr){5-7}
            & & (MAUVE) & (EM Rec.) & Rec. & Prec. & F1 & ROUGE-L & Length \\
            \midrule
            \headercolorlong
            \multicolumn{9}{c}{\textbf{Prompting-based}} \\
            \midrule
			ChatGPT & - & 73.41 & 40.37 & 72.81 & 69.69 & 71.22 & 37.92 & 39.24 \\
			\cdashlinelr{1-9}
			\multirow{3}{*}{LLaMA-2} & 7B & 79.90 & 24.32 & 17.24 & 17.87 & 17.55 & 29.38 & 42.29  \\
    		& 13B & 87.08 & 27.99 & 16.45 & 19.04 & 17.65 & 31.41 & 39.25 \\
    		& 70B & 69.28 & 31.53 & 44.18 & 44.79 & 44.48 & 31.53 & 26.86 \\
    		\cdashlinelr{1-9}
    		\multirow{3}{*}{LLaMA-2-Chat} & 7B & 66.78 & 29.93 & 55.99 & 51.66 & 53.74 & 32.93 & 26.18 \\
    		& 13B & 66.14 & 34.39 & 37.15 & 38.17 & 37.65 & 35.13 & 33.68 \\
    		& 70B & 86.60 & 41.24 & 60.19 & 61.16 & 60.67 & 37.01 & 47.09 \\
    		\cdashlinelr{1-9}
    		\multirow{2}{*}{Vicuna-v1.5} & 7B & 86.92 & 38.34 & 48.37 & 44.63 & 46.42 & 35.95 & 63.90  \\
    		& 13B & 66.11 & 35.20 & 51.92 & 53.40 & 52.65 & 35.74 & 38.57 \\
    		\cdashlinelr{1-9}
    		\multirow{2}{*}{Mistral} & 7B & 82.37 & 29.46 & 23.12 & 25.45 & 24.23 & 31.67 & 37.17  \\
    		& 8 $\times$ 7B & 83.30 & 36.30 & 32.72 & 34.49 & 33.58 & 35.05 & 38.47 \\
    		\cdashlinelr{1-9}
    		\multirow{2}{*}{Mistral-Instruct} & 7B & 82.86 & 38.57 & 64.90 & 59.67 & 62.18 & 36.21 & 45.26  \\
    		& 8 $\times$ 7B & 94.77 & 44.11 & 61.80 & 63.27 & 62.53 & 38.54 & 58.83 \\
    		\midrule
            \headercolorlong
            \multicolumn{9}{c}{\textbf{Post-hoc Retrieval}} \\
            \midrule
			ChatGPT & - & 49.78 & 37.68 & 27.11 & 27.05 & 27.08 & 36.64 & 52.61 \\
			\cdashlinelr{1-9}
			\multirow{3}{*}{LLaMA-2} & 7B & 75.56 & 16.55 & 13.88 & 13.86 & 13.87 & 26.81 & 37.50  \\
    		& 13B & 77.91 & 20.51 & 20.95 & 20.94 & 20.94 & 29.53 & 31.37 \\
    		& 70B & 75.23 & 27.58 & 28.43 & 28.43 & 28.43 & 30.33 & 29.88 \\
    		\cdashlinelr{1-9}
    		\multirow{3}{*}{LLaMA-2-Chat} & 7B & 22.50 & 14.17 & 11.33 & 11.33 & 11.33 & 21.17 & 110.04 \\
    		& 13B & 64.52 & 24.43 & 21.43 & 21.43 & 21.43 & 33.91 & 41.12 \\
    		& 70B & 70.63 & 29.68 & 24.51 & 24.51 & 24.51 & 34.17 & 45.74 \\
    		\cdashlinelr{1-9}
    		\multirow{3}{*}{Vicuna-v1.5} & 7B & 63.87 & 19.58 & 16.24 & 16.24 & 16.24 & 33.22 & 41.80  \\
    		& 13B & 73.83 & 24.79 & 24.11 & 24.11 & 24.11 & 34.42 & 43.54 \\
    		\cdashlinelr{1-9}
    		\multirow{3}{*}{Mistral} & 7B & 86.54 & 21.17 & 16.78 & 16.77 & 16.77 & 30.90 & 42.43  \\
    		& 8 $\times$ 7B & 80.99 & 36.30 & 38.37 & 35.27 & 36.75 & 35.05 & 38.47 \\
    		\cdashlinelr{1-9}
    		\multirow{3}{*}{Mistral-Instruct} & 7B & 67.97 & 26.26 & 17.87 & 17.85 & 17.86 & 33.71 & 51.56  \\
    		& 8 $\times$ 7B & 65.51 & 33.90 & 24.57 & 24.48 & 24.52 & 36.20 & 53.83 \\
    		\midrule
            \headercolorlong
            \multicolumn{9}{c}{\textbf{Training-based}} \\
            \midrule
			\multirow{2}{*}{Self-RAG} & 7B & 74.33 & 29.96 & 67.82 & 66.97 & 67.39 & 35.70 & 29.83  \\
			& 13B & 71.59 & 31.66 & 71.26 & 70.35 & 70.80 & 36.01 & 27.03  \\
			\cdashlinelr{1-9}
			\multirow{2}{*}{\vani-SFT} & 7B & 76.66 & 40.32 & 67.67 & 63.67 & 65.61 & 38.32 & 62.00  \\
			& 13B & 84.36 & 40.85 & 71.49 & 66.21 & 68.75 & 38.22 & 58.82  \\
			\cdashlinelr{1-9}
			\multirow{2}{*}{\ours} & 7B & 81.88 & 40.84 & 77.70 & 69.89 & 73.59 & 36.95 & 53.93  \\
			& 13B & 76.11 & 41.51 & 78.44 & 73.66 & 75.95 & 38.63 & 57.56  \\
            \bottomrule
        \end{tabular}
    \caption{
        ASQA full results.
    }
    \label{table:asqa_full}
\end{table*}

\begin{table*}[h]
    \centering
    \small
        \begin{tabular}{lllcccccc}
            \toprule
            \multirow{2}[4]{*}{\textbf{Model Type}} & \multirow{2}[4]{*}{\textbf{Model Size}} & \textbf{Fluency} & \textbf{Correct.} & \multicolumn{3}{c}{\textbf{Citation}}\\
            \cmidrule(lr){3-3} \cmidrule(lr){4-4} \cmidrule(lr){5-7}
            & & (MAUVE) & (Claim) & Rec. & Prec. & F1 & ROUGE-L & Length \\
            \midrule
            \headercolorlong
            \multicolumn{9}{c}{\textbf{Prompting-based}} \\
            \midrule
			ChatGPT & - & 44.65 & 12.47 & 49.44 & 47.05 & 48.22 & 20.64 & 90.2 \\
			\cdashlinelr{1-9}
			\multirow{3}{*}{LLaMA-2} & 7B & 63.72 & 4.53 & 3.92 & 5.38 & 4.54 & 18.27 & 103.36  \\
    		& 13B & 62.19 & 7.77 & 8.49 & 8.43 & 8.46 & 19.95 & 88.23 \\
    		& 70B & 53.39 & 10.43 & 23.75 & 22.43 & 23.07 & 20.43 & 93.84\\
    		\cdashlinelr{1-9}
    		\multirow{3}{*}{LLaMA-2-Chat} & 7B & 32.80 & 12.47 & 19.90 & 15.48 & 17.41 & 20.88 & 96.42  \\
    		& 13B & 29.08 & 13.83 & 16.50 & 16.09 & 16.29 & 21.04 & 94.32 \\
    		& 70B & 33.69 & 13.30 & 36.63 & 36.63 & 36.63 & 21.29 & 117.84 \\
    		\cdashlinelr{1-9}
    		\multirow{2}{*}{Vicuna-v1.5} & 7B & 31.45 & 12.30 & 29.81 & 22.45 & 25.61 & 21.36 & 105.68  \\
    		& 13B & 37.41 & 14.33 & 31.15 & 28.99 & 30.03 & 21.74 & 98.23 \\
    		\cdashlinelr{1-9}
    		\multirow{2}{*}{Mistral} & 7B & 56.62 & 8.47 & 16.04 & 16.32 & 16.18 & 20.46 & 93.80  \\
    		& 8 $\times$ 7B & 61.83 & 10.43 & 26.11 & 25.09 & 25.59 & 20.66 & 93.59 \\
    		\cdashlinelr{1-9}
    		\multirow{2}{*}{Mistral-Instruct} & 7B & 32.74 & 11.07 & 49.25 & 42.69 & 45.74 & 20.75 & 98.28  \\
    		& 8 $\times$ 7B & 38.51 & 13.93 & 49.28 & 48.34 & 48.81 & 21.34 & 113.71 \\
    		\midrule
            \headercolorlong
            \multicolumn{9}{c}{\textbf{Post-hoc Retrieval}} \\
            \midrule
			ChatGPT & - & 22.79 & 18.77 & 14.55 & 14.55 & 14.55 & 22.28 & 106.83 \\
			\cdashlinelr{1-9}
			\multirow{3}{*}{LLaMA-2} & 7B & 72.80 & 7.23 & 6.84 & 6.84 & 6.84 & 19.14 & 88.19  \\
    		& 13B & 53.21 & 10.33 & 9.61 & 9.61 & 9.61 & 20.63 & 90.44 \\
    		& 70B & 58.97 & 11.10 & 10.27 & 10.26 & 10.26 & 20.41 & 77.85 \\
    		\cdashlinelr{1-9}
    		\multirow{3}{*}{LLaMA-2-Chat} & 7B & 22.50 & 14.17 & 11.33 & 11.33 & 11.33 & 21.17 & 110.04 \\
    		& 13B & 30.36 & 14.93 & 12.10 & 12.10 & 12.10 & 21.82 & 109.79 \\
    		& 70B & 37.87 & 16.03 & 12.93 & 12.93 & 12.93 & 21.57 & 99.94 \\
    		\cdashlinelr{1-9}
    		\multirow{2}{*}{Vicuna-v1.5} & 7B & 30.88 & 11.83 & 10.91 & 10.91 & 10.91 & 21.66 & 99.03  \\
    		& 13B & 32.59 & 15.20 & 14.06 & 14.06 & 14.05 & 14.05 & 108.16 \\
    		\cdashlinelr{1-9}
    		\multirow{2}{*}{Mistral} & 7B & 52.45 & 10.47 & 8.64 & 8.64 & 8.64 & 20.48 & 90.17  \\
    		& 8 $\times$ 7B & 48.39 & 13.57 & 11.62 & 11.62 & 11.62 & 21.43 & 91.97 \\
    		\cdashlinelr{1-9}
    		\multirow{2}{*}{Mistral-Instruct} & 7B & 27.41 & 17.07 & 13.20 & 13.20 & 13.20 & 21.52 & 106.93  \\
    		& 8 $\times$ 7B & 27.60 & 17.37 & 15.68 & 15.68 & 15.68 & 21.66 & 95.21 \\
    		\midrule
            \headercolorlong
            \multicolumn{9}{c}{\textbf{Training-based}} \\
            \midrule
            \multirow{2}{*}{Self-RAG} & 7B & 30.98 & 6.90 & 22.34 & 32.40 & 26.45 & 16.48 & 41.66  \\
			& 13B & 32.04 & 6.07 & 30.46 & 40.20 & 34.66 & 15.23 & 38.19 \\
			\cdashlinelr{1-9}
			\multirow{2}{*}{\vani-SFT} & 7B & 44.12 & 9.63 & 42.30 & 40.06 & 41.15 & 20.58 & 80.43  \\
			& 13B & 46.33 & 10.27 & 46.75 & 44.47 & 45.58 & 20.56 & 84.01 \\
			\cdashlinelr{1-9}
			 \multirow{2}{*}{\ours} & 7B & 36.90 & 9.18 & 58.60 & 55.33 & 56.92 & 19.09 & 74.06  \\
			& 13B & 34.37 & 9.32 & 60.31 & 59.21 & 59.75 & 19.66 & 75.14 \\
            \bottomrule
        \end{tabular}
    \caption{
        ELI5 full results.
    }
    \label{table:eli5_full}
\end{table*}

\begin{table*}[h]
    \centering
    \small
        \begin{tabular}{lllccccc}
            \toprule
            \multirow{2}[4]{*}{\textbf{Model Type}} & \multirow{2}[4]{*}{\textbf{Model Size}} & \multicolumn{2}{c}{\textbf{Correctness}} & \multicolumn{3}{c}{\textbf{Citation} }\\
            \cmidrule(lr){3-4} \cmidrule(lr){5-7}
            & & Rec.-5 & Prec. & Rec. & Prec. & F1 & Num Pred. \\
            \midrule
            \headercolorlong
            \multicolumn{8}{c}{\textbf{Prompting-based}} \\
            \midrule
			ChatGPT & - & 20.28 & 19.84 & 19.06 & 22.03 & 20.44 & 4.71 \\
			\cdashlinelr{1-8}
			\multirow{3}{*}{LLaMA-2} & 7B & 12.56 & 11.32 & 6.03 & 6.35 & 6.19 & 7.02  \\
    		& 13B & 18.00 & 12.39 & 5.45 & 5.74 & 5.59 & 11.31 \\
    		& 70B & 18.50 & 14.79 & 10.10 & 10.50 & 10.30 & 8.31 \\
    		\cdashlinelr{1-8}
			\multirow{3}{*}{LLaMA-2-Chat} & 7B & 17.96 & 19.74 & 9.58 & 9.68 & 9.63 & 4.73  \\
    		& 13B & 21.34 & 18.86 & 8.94 & 9.06 & 9.00 & 6.51 \\
    		& 70B & 22.62 & 18.04 & 13.49 & 13.98 & 13.73 & 7.44 \\
    		\cdashlinelr{1-8}
			\multirow{2}{*}{Vicuna-v1.5} & 7B & 14.22 & 14.74 & 11.26 & 11.64 & 11.45 & 5.87  \\
    		& 13B & 22.06 & 19.60 & 13.04 & 13.74 & 13.38 & 7.62 \\
    		\cdashlinelr{1-8}
    		\multirow{2}{*}{Mistral} & 7B & 16.96 & 15.98 & 7.50 & 7.76 & 7.63 & 6.29 \\
    		& 8 $\times$ 7B & 18.18 & 15.63 & 9.72 & 10.20 & 9.95 & 6.63 \\
    		\cdashlinelr{1-8}
			\multirow{2}{*}{Mistral-Instruct} & 7B & 17.52 & 21.29 & 17.56 & 18.53 & 18.03 & 4.54 \\
    		& 8 $\times$ 7B & 20.12 & 19.64 & 19.27 & 20.38 & 19.81 & 5.32 \\
    		\midrule
    		\headercolorlong
    		\multicolumn{8}{c}{\textbf{Post-hoc Retrieval}} \\
            \midrule
			ChatGPT & - & 25.14 & 22.85 & 12.29 & 12.29 & 12.29 & 5.46 \\
			\cdashlinelr{1-8}
			\multirow{3}{*}{LLaMA-2} & 7B & 6.48 & 5.11 & 5.05 & 5.05 & 5.05 & 6.55  \\
    		& 13B & 9.88 & 7.17 & 5.20 & 5.20 & 5.20 & 6.98 \\
    		& 70B & 14.44 & 12.44 & 7.49 & 7.49 & 7.49 & 7.41 \\
    		\cdashlinelr{1-8}
			\multirow{3}{*}{LLaMA-2-Chat} & 7B & 12.94 & 10.89 & 7.76 & 7.76 & 7.76 & 5.99  \\
    		& 13B & 15.72 & 12.23 & 7.87 & 7.87 & 7.87 & 6.32 \\
    		& 70B & 17.90 & 14.45 & 9.05 & 9.05 & 9.05 & 6.05 \\
    		\cdashlinelr{1-8}
			\multirow{2}{*}{Vicuna-v1.5} & 7B & 12.04 & 9.71 & 6.69 & 6.69 & 6.69 & 7.10  \\
    		& 13B & 14.78 & 11.47 & 8.50 & 8.50 & 8.50 & 6.67 \\
    		\cdashlinelr{1-8}
    		\multirow{2}{*}{Mistral} & 7B & 9.94 & 7.90 & 6.00 & 6.00 & 6.00 & 7.38  \\
    		& 8 $\times$ 7B & 13.92 & 12.08 & 6.70 & 6.70 & 6.70 & 6.58 \\
    		\cdashlinelr{1-8}
			\multirow{2}{*}{Mistral-Instruct} & 7B & 15.80 & 12.15 & 8.34 & 8.34 & 8.34 & 7.01  \\
    		& 8 $\times$ 7B & 24.16 & 18.28 & 9.78 & 9.78 & 9.78 & 7.37 \\
    		\midrule
    		\headercolorlong
    		\multicolumn{8}{c}{\textbf{Training-based}} \\
    		\midrule
    		\multirow{2}{*}{Self-RAG} & 7B & 2.34 & 1.98 & 10.53 & 18.80 & 13.50 & 3.49  \\
    		& 13B & 1.90 & 1.33 & 12.79 & 20.90 & 15.86 & 3.08 \\
    		\cdashlinelr{1-8}
    		\multirow{2}{*}{\vani-SFT} & 7B & 12.86 & 21.09 & 21.35 & 21.36 & 21.35 & 7.49  \\
    		& 13B & 12.68 & 22.80 & 23.64 & 23.71 & 23.67 & 3.14 \\
    		\cdashlinelr{1-8}
    		\multirow{2}{*}{\ours} & 7B & 11.50 & 21.38 & 24.74 & 24.84 & 24.79 & 3.08  \\
    		& 13B & 11.94 & 22.61 & 24.86 & 25.39 & 25.12 & 3.17 \\
    	
            \bottomrule
        \end{tabular}
    \caption{
        QAMPARI full results.
    }
    \label{table:qampari_full}
\end{table*}

\end{document}